\documentclass[times,twocolumn,final]{elsarticle}

\usepackage{medima}
\usepackage{framed,multirow}

\usepackage{amssymb}
\usepackage{latexsym}
\usepackage{graphicx}
\usepackage{makecell}
\usepackage{array}
\usepackage{mdwmath}
\usepackage{mdwtab}
\usepackage{eqparbox}
\usepackage{algorithmic}
\usepackage{algorithm}
\usepackage{amsmath}
\usepackage{url}
\usepackage{xcolor}
\usepackage{multirow}
\usepackage{hyperref}
\usepackage{cleveref}

\definecolor{newcolor}{rgb}{.8,.349,.1}

\journal{Medical Image Analysis}

\begin{document}

\verso{Yichi Zhang \textit{et~al.}}

\begin{frontmatter}

\title{SemiSAM-O1: Pushing the Boundary of Annotation-Efficient Medical Image Segmentation with Generalist Knowledge Fusion}%

\author[1,2]{Yichi \snm{Zhang}}
\author[2]{Le \snm{Xue}}
\author[3]{Bichun \snm{Xu}}
\author[3]{Judong \snm{Luo}}
\author[4]{Zhigang \snm{Wu}}
\author[4]{Yu \snm{Fu}}
\author[1,2]{Zixin \snm{Hu}}
\author[1,2,6]{Yuan \snm{Cheng}}
\author[1,2,5,6]{Yuan \snm{Qi}}

\address[1]{Artificial Intelligence Innovation and Incubation Institute, Fudan University, Shanghai, China.}
\address[2]{Shanghai Academy of Artificial Intelligence for Science, Shanghai, China.}
\address[3]{Department of Radiotherapy, Tongji Hospital, School of Medicine, Tongji University, Shanghai, China.}
\address[4]{School of Information Science and Engineering, Lanzhou University, Lanzhou, China.}
\address[5]{Department of Information and Intelligence Development, Zhongshan Hospital, Fudan University, Shanghai, China.}
\address[6]{Shanghai Innovation Institute, Shanghai, China.}

\received{}
\finalform{}
\accepted{}
\availableonline{}

\begin{abstract}
Semi-supervised learning (SSL) has become a promising solution to alleviate the annotation burden of deep learning-based medical image segmentation models.
While recent advances in foundation model-driven SSL have pushed the boundary to extremely limited annotation scenarios, they fail to maintain robust competitive performance in complex imaging modalities.
In this paper, we propose \textbf{SemiSAM-O1}, an annotation-efficient framework using \textbf{only one} annotated template case for segmentation.
SemiSAM-O1 extends the specialist-generalist collaborative learning framework to the extreme one-label setting by fully exploiting the foundation model's feature representation capability beyond its prompting interface.
SemiSAM-O1 operates in two stages. In the first stage, the foundation model's encoder extracts dense features from all volumes, and class prototypes derived from the single annotated template are propagated to the unlabeled pool via feature similarity to produce coarse initial pseudo-labels. In the second stage, an iterative training-and-refinement loop progressively improves both the segmentation model and the pseudo-labels over multiple rounds, where each round trains the model from scratch on current pseudo-labels and generates updated predictions with voxel-wise uncertainty estimates. An uncertainty-guided refinement step further leverages the foundation model's global feature space to correct high-uncertainty regions by aggregating labels from their most similar confident neighbors, establishing a virtuous cycle of mutual improvement.
Extensive experiments on a wide range of segmentation tasks across different modalities and anatomical targets demonstrate that SemiSAM-O1 significantly narrows the performance gap between one-label semi-supervised learning and full supervision, while significantly reducing the computational overhead of online foundation model inference.
Our code is available at \url{https://github.com/YichiZhang98/SemiSAM-O1}.
\end{abstract}


\end{frontmatter}


\begin{figure*}[t]
\centering
\includegraphics[width=\linewidth]{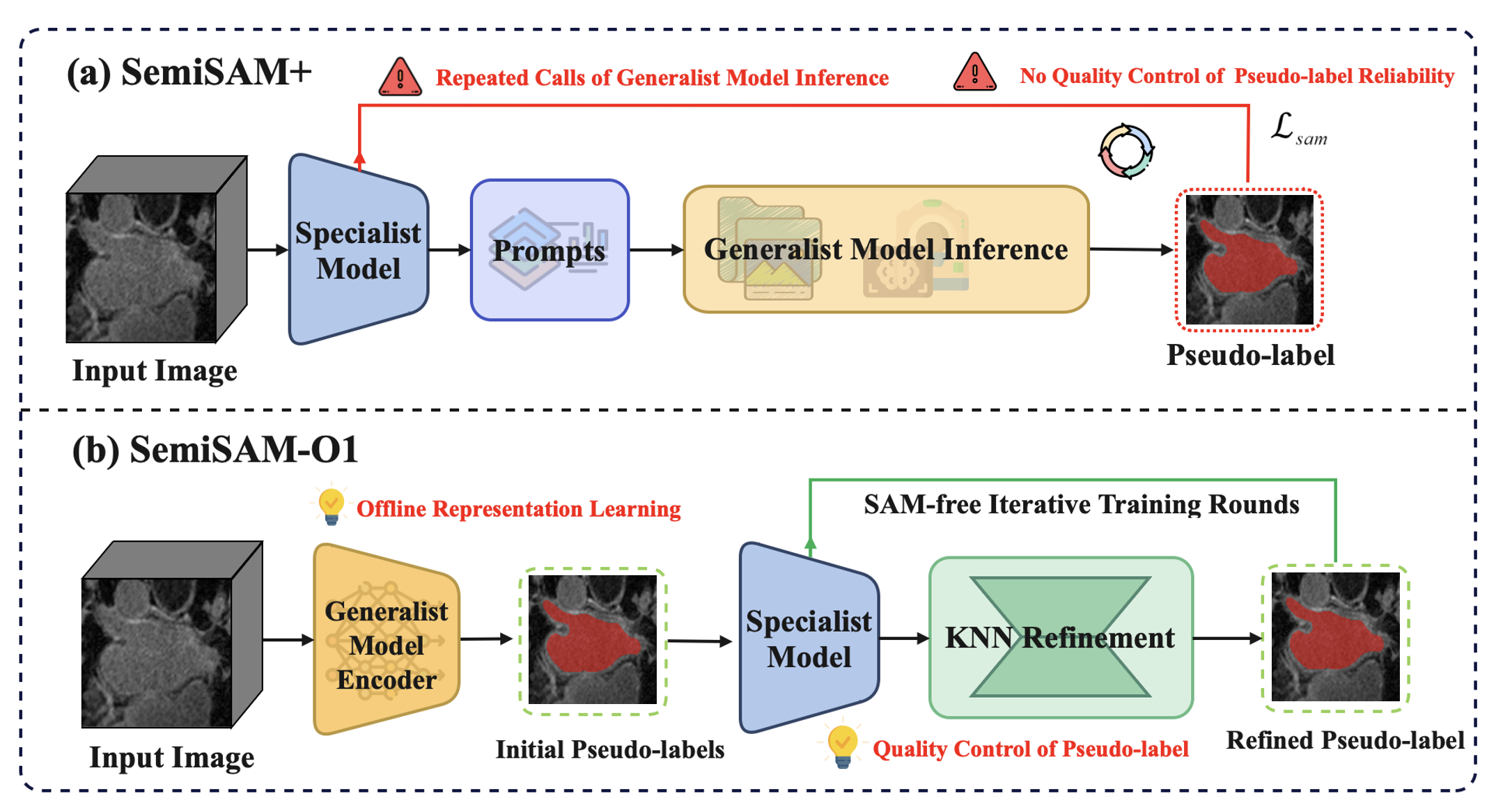}
\caption{Comparison of SemiSAM+ and the proposed SemiSAM-O1. SemiSAM+ repeatedly invokes SAM inference during training to compute a consistency loss between the specialist model and SAM outputs, incurring substantial computational overhead with no quality control on SAM's output reliability. In contrast, SemiSAM-O1 employs SAM only once in an offline stage to learn representations for prototype-based pseudo-label initialization, after which all iterative rounds of specialist training, pseudo-label generation and refinement proceed without SAM involvement.}
\label{fig_comparison}
\end{figure*}

\section{Introduction}

Medical image segmentation aims to precisely delineate specific anatomical regions and pathological abnormalities from various medical imaging, which is an essential step for accurate disease diagnosis and treatment planning \citep{MSD,lalande2021deep,AbdomenCT-1K,zhang2024nasalseg,xu2025multicenter}.
While deep learning has delivered exceptional performance in a broad spectrum of organ and lesion segmentation tasks, its inherent dependence on large-scale high-quality annotated datasets for model training persists as a major bottleneck. This challenge is especially pronounced in the medical domain, where the generation of expert-level annotations is both prohibitively expensive and extremely labor-intensive \citep{tajbakhsh2020embracing}.
In real-world clinical settings, unlabeled data are inherently widely available and abundant. Against this background, semi-supervised learning (SSL) has emerged as a highly promising and practical paradigm, which enables the efficient joint utilization of scarce labeled data and large-scale unlabeled data \citep{SemiSurvey}.

In our view, the success of SSL for medical image segmentation is fundamentally determined by two core criteria. First, the model must be capable of rapidly capturing sufficient discriminative representations from a limited set of labeled samples. Second, equipped with such discriminative representations, the model should effectively exploit unlabeled data for performance optimization. As highlighted in our prior work SemiSAM+ \citep{SemiSAM+}, existing SSL methodologies predominantly concentrate on the second criterion, with most strategies designed to improve the utilization efficiency of unlabeled data under the premise of a relatively large annotated dataset (e.g. 10\% or 20\%). Nevertheless, these methods suffer from severe performance degradation in extremely data-scarce settings as they cannot learn adequate discriminative information from the severely limited labeled data. To address this critical limitation, SemiSAM+ leverages SAM-like general-purpose promptable segmentation foundation models, which are pre-trained on large-scale data with learned transferable representations and enable highly efficient interactive segmentation through prompting. By introducing a novel paradigm for specialist–generalist collaborative learning, SemiSAM+ achieves significant performance improvement under limited annotation scenarios.

However, as illustrated in Fig.~\ref{fig_comparison}, SemiSAM+ repeatedly invokes SAM inference during training to enforce specialist–generalist consistency, which introduces substantial computational overhead. Besides, SemiSAM+ lacks any quality control on the reliability of SAM's outputs, where noisy or incorrect SAM predictions are directly used to supervise the specialist model, potentially reinforcing errors. As a result, a pronounced performance gap persists compared with the fully supervised performance.

It is worth noting that human clinicians inherently possess one-shot generalization ability. When given a single annotated example, they can delineate corresponding target regions with favorable segmentation performance, without relying on large-scale annotated data. Such strong one-shot learning capability inspires us to rethink existing semi-supervised paradigms, revealing that conventional SSL settings built upon moderate labeled samples are far from the ultimate limit of low-annotation segmentation. Therefore, in this work, we aim to push the boundaries of annotation efficiency by tackling the extreme challenge of training a segmentation model with just a single annotated image and a pool of unlabeled data.

In this work, we propose SemiSAM-O1, which extends the specialist-generalist collaborative learning framework to address the challenging one-label semi-supervised segmentation scenario. Different from SemiSAM+, SemiSAM-O1 introduces an iterative pseudo-label refinement pipeline that unlocks the foundation model's feature representation capability beyond its prompting interface.
Specifically, given only a single annotated volume, SemiSAM-O1 employs the generalist foundation model \emph{only once} in an offline stage to extract encoder features for prototype-based pseudo-label initialization, after which all iterative training rounds proceed without repeated calls of generalist model inference. It then enters a multi-round loop: in each round, a specialist model is trained from scratch on the current pseudo-labeled data; the trained model generates updated pseudo-labels with uncertainty estimates; and an uncertainty-guided $K$-nearest-neighbor refinement step corrects high-uncertainty predictions by querying their most similar confident neighbors in the foundation model's feature space. This design offers an explicit quality control mechanism absent in SemiSAM+, ensuring that only reliable predictions contribute to pseudo-label updates. The iterative process creates a virtuous cycle where improved pseudo-labels lead to better-trained models, which in turn produce more accurate pseudo-labels.
Our contributions can be summarized as follows.

\begin{itemize}
\item[-] We propose SemiSAM-O1, which extends existing semi-supervised segmentation framework to the extreme one-label setting by introducing an iterative pseudo-label refinement pipeline that exploits the foundation model as an offline feature extractor for annotation propagation and pseudo-label quality control, without requiring any online generalist inference during training.
\item[-] We design a prototype-based initialization strategy that leverages the foundation model's dense feature representations to propagate annotations from a single labeled volume to the entire unlabeled pool, and an uncertainty-guided KNN refinement mechanism that progressively corrects erroneous pseudo-labels across iterative rounds using inter-sample relationships in the foundation model's feature space.
\item[-] Extensive experiments across multiple semi-supervised learning backbones on a wide range of segmentation tasks across different modalities and anatomical targets demonstrate that SemiSAM-O1 substantially narrows the performance gap between one-label semi-supervised learning and full supervision, while significantly reducing the training time required for online inference.
\end{itemize}

\section{Related Work}

\subsection{Conventional Semi-Supervised Medical Image Segmentation}

A direct and intuitive paradigm for leveraging unlabeled data in semi-supervised medical image segmentation involves assigning pseudo-annotations to unlabeled images, which are then used alongside labeled data to update the segmentation model \citep{lee2013pseudo}. Existing approaches under this framework primarily focus on designing advanced strategies for pseudo-label generation \citep{seibold2021reference} and quality selection \citep{huang2022mtl,wang2022ssa}, aiming to efficiently utilize pseudo-labeled samples to guide the model learning process.
Beyond the iterative paradigm of pseudo-label generation and model update, recent advancements in semi-supervised learning (SSL) have shifted toward enhancing model representation learning by introducing unsupervised regularization signals, enabling more effective exploitation of large-scale unlabeled data. One major line of research in this direction focuses on generating prediction discrepancies under various perturbations: representative methods include self-ensembling \citep{tarvainen2017mean,yu2019uncertainty}, perturbations of input resolution scales \citep{luo2022semi}, and data recombination techniques \citep{chen2023magicnet}.

Another branch of research avoids data perturbation for consistency learning and instead constructs task-level regularization. Specific strategies include adversarial training \citep{zhang2017deep,xie2023adversarial}, integrating auxiliary tasks to leverage geometric information of anatomical structures \citep{li2020shape,luo2021semi,chen2022semi,zhang2023uncertainty}, and exploiting diverse learning procedures of different backbone networks \citep{luo2022semi,ma2024semi,song2024sdcl}.
In summary, conventional SSL methods for medical image segmentation are predominantly model-centric, focusing on designing innovative strategies to generate unsupervised supervision signals embedded in unlabeled data. Their learning process heavily relies on transferring knowledge from labeled images to unlabeled ones. However, a critical limitation of these approaches is their reliance on a certain annotation budget: when applied to new segmentation tasks, they still require annotating a small subset of the dataset to achieve acceptable performance, and struggle to deliver satisfactory results in scenarios with extremely limited annotations.

\begin{figure*}[t]
\centering
\includegraphics[width=\linewidth]{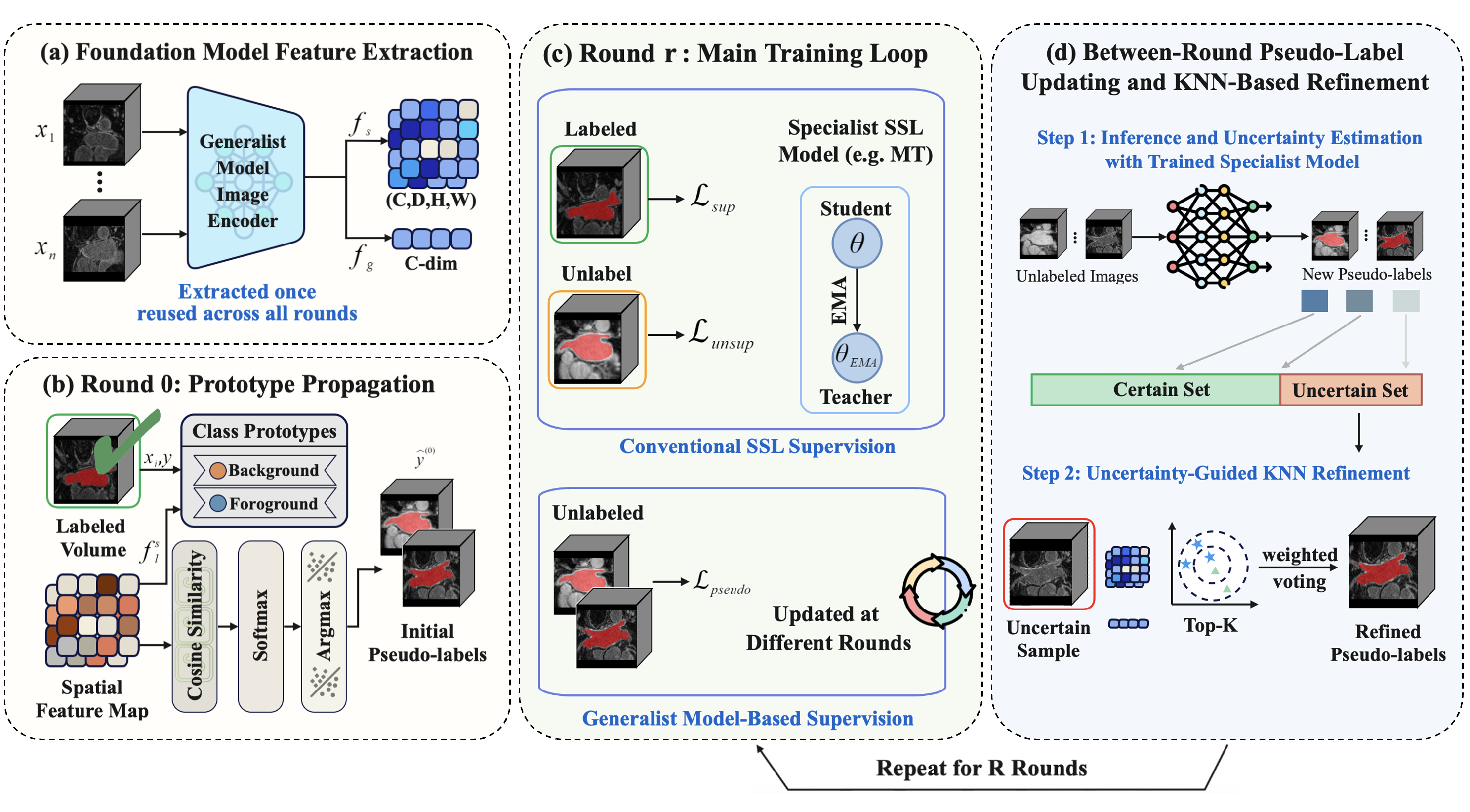}
\caption{Overview of the proposed SemiSAM-O1 framework. (a)~The generalist foundation model encoder extracts spatial and global features from all volumes once before training. (b)~In Round~0, class prototypes computed from a single labeled volume are propagated to the unlabeled pool via cosine similarity to produce initial pseudo-labels. (c)~In each subsequent round, a specialist SSL model is trained under conventional semi-supervised supervision on the real label and generalist model-based supervision on pseudo-labels that are updated between rounds. (d)~After training, the specialist generates new pseudo-labels with uncertainty estimates, and an uncertainty-guided KNN refinement step corrects low-confidence predictions by querying their nearest neighbors in the foundation model's feature space. Components (c) and (d) repeat for R rounds to progressively improve pseudo-label quality and model performance.}
\label{fig_framework}
\end{figure*}

\subsection{Foundation Models for Promptable Segmentation}

Foundation models pre-trained on large-scale datasets have recently attracted increasing attention in the field of medical imaging \citep{willemink2022toward,moor2023foundation,zhang2024challenges}.
In image segmentation, the Segment Anything Model (SAM) \citep{SAM} introduces a novel paradigm of promptable segmentation, enabling interactive extraction of arbitrary targets using positional prompts such as points and bounding boxes.
Trained on large-scale datasets, SAM learns highly generalizable visual representations and demonstrates remarkable zero-shot generalization across diverse segmentation tasks \citep{SAM4MIS}.
However, a significant domain gap exists between natural images and medical images, which limits the direct application of SAM in clinical scenarios.
To bridge this gap, a series of medical adaptations have been proposed \citep{MedSAM,SAM-Med2D,SAM-Med3D,du2024segvol,nninteractive}.
Beyond their direct use for interactive segmentation, these foundation models encode rich pre-trained representations, which capture semantic and structural information transferable to downstream tasks.

\subsection{Foundation Model-Based Semi-Supervised Medical Image Segmentation}

The emergence of promptable segmentation foundation models has opened a new avenue for semi-supervised medical image segmentation by providing an additional source of supervision beyond the conventional labeled-unlabeled paradigm \citep{zhang2026gsfm}.
As one of the earliest attempts in this direction, SemiSAM \citep{zhang2024semisam} unlocks the potential of SAM by introducing an extra supervision branch that enforces consistency between the specialist model's predictions and SAM's prompted outputs on unlabeled data.
SemiSAM+ \citep{SemiSAM+} delivers substantial methodological improvements by adopting more diverse prompt formulations including mask-based, point-based, and uncertainty-aware prompting strategies, and conducts extensive validations upon different specialist and generalist segmentation models.
Building upon this specialist-generalist collaborative paradigm, numerous subsequent studies have further optimized the utilization of generalist models \citep{li2025foundation,li2025stitching,jiang2025foundation,miao2026sam}.
Despite the significant improvement brought by foundation model-driven regularization, these methods still rely on a 
minimum annotation budget and experience notable performance degradation under the extreme one-label setting. The core limitation lies in that the specialist model cannot learn sufficient discriminative features to generate meaningful prompts for the generalist model in the early stage of training, limiting the effectiveness of the specialist generalist collaborative loop.

\section{Methodology}

\subsection{Preliminaries} \label{sec:prelim}

Semi-supervised learning aims to utilize unlabeled data in conjunction with labeled data to train higher-performing segmentation models with limited annotations.
Given a dataset $\mathcal{D}$ for training, we annotate a small subset with $M$ cases as $\mathcal{D}_{L} = \{x_{l}^{i}, y^{i}\}_{i=1}^{M}$ based on the available budget, while the remaining unlabeled set with $N$ unlabeled cases as $\mathcal{D}_{U} = \{x_{u}^{i}\}_{i=1}^{N}$, where $x_{l}$ and $x_{u}$ denote the labeled and unlabeled input images, and $y$ denotes the corresponding ground truth of labeled data. Generally, $\mathcal{D}_{L}$ is a relatively small subset of the entire dataset $\mathcal{D}$, which means $M \ll N$.
In this work, we focus on the extreme \textbf{one-label} setting where $M=1$, i.e., only a single annotated volume is available for supervision.
The model is optimized based on the combination of supervised segmentation loss $\mathcal{L}_{sup}(\theta;\mathcal{D}_{L})$ and unsupervised regularization loss $\mathcal{L}_{unsup}(\theta;\mathcal{D}_{U})$:
\begin{equation}
\min \limits_{\theta} \mathcal{L}_{sup}(\theta;\mathcal{D}_{L}) + \lambda \mathcal{L}_{unsup}(\theta;\mathcal{D}_{U})
\label{loss_ssl}
\end{equation}
For the SAM-like generalist foundation model, the model adopts an image encoder $\mathbf{I}$, a prompt encoder $\mathbf{P}$, and a mask decoder $\mathbf{M}$ to perform promptable segmentation:
\begin{equation}
F_{\Theta}(x_{i},p(x_{i}) ) = \mathbf{M}[ \mathbf{I}(x_{i}) + \mathbf{P}(p(x_{i})) ]
\label{eq_sam}
\end{equation}
where $p(x_{i})$ represents the positional prompt for input image $x_{i}$. In this work, we use SAM-Med3D~\citep{SAM-Med3D}, a 3D medical adaptation of SAM, as the default generalist foundation model.

\subsection{Recap of SemiSAM+} \label{sec:semisam_recap}

SemiSAM+~\citep{SemiSAM+} introduces a specialist-generalist collaborative learning framework for semi-supervised medical image segmentation. The key idea is to leverage a frozen generalist foundation model to provide additional supervision for the trainable specialist model.
During training, the specialist model's coarse prediction is used to generate positional prompts $p_{n}(x_{i}) = \text{Prompting}_{n} ( S_{\theta}(x_{i}) )$ for the generalist model. The generalist model then produces segmentation outputs based on these prompts. By enforcing consistency between the specialist and generalist outputs through a confidence-aware regularization, SemiSAM+ effectively transfers the pre-trained knowledge of the foundation model to guide the specialist model's learning. The uncertainty of the generalist output is estimated via prediction discrepancy across different prompt types:
\begin{equation}
U_{x} = \mathit{D} [ F_{\Theta}(x,p_{1}(x) ) , \ldots , F_{\Theta}(x,p_{n}(x))]
\label{eq_unc_sam}
\end{equation}
The overall training loss of SemiSAM+ is:
\begin{equation}
\min \limits_{\theta} \mathcal{L}_{sup}(\theta;\mathcal{D}_{L}) + \lambda \mathcal{L}_{unsup}(\theta;\mathcal{D}_{U}) + \beta \mathcal{L}_{sam}(\theta;\Theta;\mathcal{D}_{U})
\label{loss_semisam}
\end{equation}
where $\mathcal{L}_{sam}$ denotes the confidence-aware consistency regularization between the specialist and generalist outputs.

While SemiSAM+ demonstrates significant improvements over conventional SSL methods, a pronounced performance gap persists compared with the fully supervised performance under the one-label setting. With only a single labeled case, the specialist model cannot learn sufficient discriminative features to generate meaningful prompts for the generalist model in the early stage of training, limiting the effectiveness of the specialist-generalist collaborative loop.
In addition, the online inference of the generalist model required in each training iteration leads to low efficiency and a substantial increase in computational overhead. These limitations motivate a new design that leverages the foundation model's pre-trained features offline, rather than coupling it into the training loop.

\begin{table*}[t]
	\caption{Comparison of FS Baseline, SSL Baseline, SemiSAM+, and SemiSAM-O1 on left atrium segmentation under the one-label setting ($M=1$). All specialist models use 3D U-Net as the backbone. } \label{Table_oneseg}
	\centering
    \normalsize
	\renewcommand\arraystretch{1.35}
	\begin{tabular}{c|c|cccc}
		\hline \hline
		SSL Backbone & Method & Dice$\uparrow$[\%] & Jaccard$\uparrow$[\%] & 95HD$\downarrow$[voxel] & ASD$\downarrow$[voxel] \\ \hline
		-- & FS Baseline & 37.97$\pm$21.80 & 25.83$\pm$17.98 & 39.50$\pm$13.57 & 16.46$\pm$6.72 \\ \hline
		\multirow{5}{*}{MT} & SSL Baseline & 40.64$\pm$23.82 & 28.42$\pm$19.77 & 38.92$\pm$14.64 & 15.85$\pm$7.19 \\
		& SemiSAM+ & 48.91$\pm$22.79 & 35.46$\pm$20.77 & 38.49$\pm$15.40 & 14.61$\pm$7.12 \\
		& SemiSAM-O1 (R1) & 72.02$\pm$10.42 & 57.27$\pm$12.28 & 28.25$\pm$13.43 & 8.87$\pm$4.45 \\
		& SemiSAM-O1 (R2) & \textbf{76.78$\pm$8.02} & \textbf{62.96$\pm$9.96} & \textbf{20.24$\pm$9.83} & \textbf{5.84$\pm$3.00} \\
		& SemiSAM-O1 (R3) & 75.66$\pm$9.56 & 61.73$\pm$11.64 & 21.21$\pm$11.12 & 6.63$\pm$4.18 \\ \hline
		\multirow{5}{*}{UA-MT} & SSL Baseline & 41.73$\pm$23.84 & 29.46$\pm$20.87 & 39.74$\pm$16.72 & 15.81$\pm$8.20 \\
		& SemiSAM+ & 50.10$\pm$22.83 & 36.56$\pm$20.87 & 37.02$\pm$15.08 & 14.27$\pm$7.16 \\
		& SemiSAM-O1 (R1) & 70.03$\pm$11.30 & 55.05$\pm$13.46 & 25.26$\pm$9.76 & 8.98$\pm$4.09 \\
		& SemiSAM-O1 (R2) & 71.64$\pm$10.61 & 56.83$\pm$12.41 & 32.10$\pm$14.37 & 9.96$\pm$4.80 \\
		& SemiSAM-O1 (R3) & \textbf{75.60$\pm$8.26} & \textbf{61.46$\pm$10.37} & \textbf{24.75$\pm$11.49} & \textbf{7.70$\pm$4.31} \\ \hline
		\multirow{5}{*}{DAN} & SSL Baseline & 44.61$\pm$20.52 & 31.08$\pm$18.10 & 38.03$\pm$14.49 & 14.68$\pm$6.31 \\
		& SemiSAM+ & 62.26$\pm$14.34 & 46.82$\pm$15.67 & 31.88$\pm$10.48 & 11.94$\pm$5.07 \\
		& SemiSAM-O1 (R1) & 69.32$\pm$11.29 & 54.21$\pm$13.46 & 27.78$\pm$11.10 & 9.70$\pm$4.28 \\
		& SemiSAM-O1 (R2) & 71.14$\pm$9.12 & 55.97$\pm$10.81 & 24.29$\pm$11.38 & 8.18$\pm$4.59 \\
		& SemiSAM-O1 (R3) & \textbf{73.39$\pm$11.36} & \textbf{59.19$\pm$13.79} & \textbf{21.81$\pm$10.15} & \textbf{7.15$\pm$4.41} \\ \hline
		\multirow{5}{*}{DTC} & SSL Baseline & 41.62$\pm$21.49 & 28.76$\pm$18.51 & 38.08$\pm$14.71 & 15.02$\pm$7.18 \\
		& SemiSAM+ & 53.52$\pm$23.05 & 39.84$\pm$21.25 & 35.19$\pm$17.46 & 13.25$\pm$8.61 \\
		& SemiSAM-O1 (R1) & 72.71$\pm$10.94 & 58.25$\pm$13.13 & 28.34$\pm$12.28 & 9.38$\pm$4.67 \\
		& SemiSAM-O1 (R2) & 73.18$\pm$9.93 & 58.65$\pm$12.11 & \textbf{21.39$\pm$11.85} & \textbf{6.20$\pm$4.92} \\
		& SemiSAM-O1 (R3) & \textbf{73.83$\pm$9.35} & \textbf{59.37$\pm$11.51} & 22.33$\pm$11.39 & 6.22$\pm$4.74 \\ \hline
		-- & FS Upperbound & 88.18$\pm$5.30 & 79.23$\pm$7.71 & 11.34$\pm$11.85 & 2.90$\pm$3.27 \\ \hline \hline
	\end{tabular}
\end{table*}

\subsection{SemiSAM-O1: Learning with Only One Annotated Image} \label{sec:semisam_o1}

To address the challenge of learning from a single labeled case, SemiSAM-O1 introduces a multi-round iterative pipeline that exploits the feature representation capability of the foundation model to progressively improve pseudo-labels for unlabeled data.
An overview of the proposed pipeline is illustrated in Fig.~\ref{fig_framework}. The pipeline consists of four components: (a)~a one-time foundation model feature extraction stage, (b)~an initialization phase (Round~0) that generates pseudo-labels via prototype propagation, (c)~a main training loop where a specialist SSL model is trained with the current pseudo-labels, and (d)~a between-round pseudo-label update and KNN-based refinement step. Components (c) and (d) are repeated for $R$ rounds. We describe each component below.

\textbf{Foundation Model Feature Extraction (Fig.~\ref{fig_framework}(a)).}
Beyond its role as a promptable segmentor, the foundation model's image encoder $\mathbf{I}$ provides a powerful feature extractor pre-trained on large-scale datasets. For each volume $x_i \in \mathcal{D}$, we extract spatial feature maps and a global feature vector using the encoder:
\begin{equation}
\mathbf{f}^{s}_{i} = \mathbf{I}(x_i) \in \mathbb{R}^{C \times D' \times H' \times W'}
\end{equation}
\begin{equation}
\mathbf{f}^{g}_{i} = \frac{\text{GAP}(\mathbf{f}^{s}_{i})}{\| \text{GAP}(\mathbf{f}^{s}_{i}) \|_2} \in \mathbb{R}^{C}
\label{eq:global_feat}
\end{equation}
where $\text{GAP}(\cdot)$ denotes global average pooling, and $(D', H', W')$ is the spatial resolution of the feature map determined by the encoder's patch size.
The spatial features $\mathbf{f}^{s}_{i}$ capture local structural information at each patch location, while the global features $\mathbf{f}^{g}_{i}$ provide a holistic representation of each volume for cross-sample similarity comparison. These features are extracted once before training and reused across all rounds.

\textbf{Round 0: Initial Pseudo-Label Generation via Prototype Propagation (Fig.~\ref{fig_framework}(b)).}
Given only a single labeled volume $(x_l, y)$, Round~0 generates initial pseudo-labels $\hat{\mathcal{Y}}^{(0)} = \{\hat{y}^{(0)}_u\}_{u=1}^{N}$ for all unlabeled data by propagating the annotation through feature similarity.
We first compute a class prototype for each semantic class $c \in \{0, \ldots, C_{\text{cls}}-1\}$ by aggregating feature vectors from the labeled volume at spatial locations corresponding to class $c$. The ground truth $y$ is downsampled to the feature resolution via nearest-neighbor interpolation, yielding $y' \in \{0, \ldots, C_{\text{cls}}-1\}^{D' \times H' \times W'}$. The prototype for class $c$ is:
\begin{equation}
\mathbf{p}_c = \text{Normalize}\left(\frac{\sum_{(d,h,w)} \mathbb{I}(y'[d,h,w] = c) \cdot \mathbf{f}^{s}_{l}[:, d, h, w]}{\sum_{(d,h,w)} \mathbb{I}(y'[d,h,w] = c) + \epsilon}\right)
\label{eq:prototype}
\end{equation}
where $\mathbb{I}(\cdot)$ is the indicator function, $\epsilon$ is a small constant for numerical stability, and $\text{Normalize}(\cdot)$ denotes L2 normalization.

For each unlabeled volume $x_u$, we then compute the cosine similarity between its normalized spatial features and each class prototype to produce class-wise similarity maps $\mathcal{S}_{u,c} \in \mathbb{R}^{D' \times H' \times W'}$. The low-resolution similarity maps are upsampled to the original volume resolution via nearest-neighbor upsampling, yielding $\tilde{\mathcal{S}}_{u,c}$. The initial hard pseudo-label is generated by applying a pixel-wise softmax across classes followed by argmax:
\begin{equation}
\hat{y}^{(0)}_u[v] = \arg\max_{c} \frac{\exp(\tilde{\mathcal{S}}_{u,c}[v])}{\sum_{j} \exp(\tilde{\mathcal{S}}_{u,j}[v])}
\label{eq:pseudo_r0}
\end{equation}
These prototype-propagated pseudo-labels provide a coarse but reasonable initialization for the unlabeled set by leveraging the foundation model's ability to capture semantic similarity across volumes from a single reference annotation.

\textbf{Rounds 1 to $R$: Iterative Model Training and Pseudo-Label Refinement.}
For rounds $r = 1, \ldots, R$, the pipeline iteratively trains a specialist model with the current pseudo-labels, generates improved pseudo-labels from the trained model, and refines uncertain predictions using the foundation model's features. Each round consists of three stages.

In the first stage (Fig.~\ref{fig_framework}(c)), a specialist SSL model $S^{(r)}_{\theta}$ and its corresponding EMA teacher model are initialized from scratch. SemiSAM-O1 is agnostic to the choice of SSL backbone and can be instantiated with various methods such as Mean Teacher~\citep{tarvainen2017mean} or UAMT~\citep{yu2019uncertainty}. The training set combines the single labeled volume with all unlabeled volumes annotated by the pseudo-labels $\hat{\mathcal{Y}}^{(r-1)}$ from the previous round. As shown in Fig.~\ref{fig_framework}(c), the training loop receives two sources of supervision: conventional SSL supervision ($\mathcal{L}_{sup}$ on the real label, $\mathcal{L}_{unsup}$ on unlabeled data via student-teacher consistency) and generalist model-based supervision ($\mathcal{L}_{pseudo}$ on pseudo-labels that are updated between rounds). The total training loss at each iteration is:
\begin{equation}
\mathcal{L}^{(r)} = \mathcal{L}_{sup} + \lambda \mathcal{L}_{unsup} + \alpha^{(r)} \mathcal{L}_{pseudo}
\label{eq:round_loss}
\end{equation}
where $\mathcal{L}_{pseudo} = \frac{1}{2}(\mathcal{L}_{ce}(S_\theta(x_u), \hat{y}_u) + \mathcal{L}_{dice}(S_\theta(x_u), \hat{y}_u))$ is the supervised loss on pseudo-labeled data, and $\alpha^{(r)}$ is a ramp-up weight that linearly increases from 0 to 1 during the first 30\% of training iterations to avoid early fitting to noisy pseudo-labels.

Unlike SemiSAM+ which requires online generalist inference at each training iteration (Eq.~\ref{loss_semisam}), SemiSAM-O1 completely decouples the foundation model from the training loop, using it only for offline feature extraction and between-round pseudo-label refinement. Re-initialization at each round prevents the model from inheriting biased representations learned from potentially inaccurate pseudo-labels in earlier rounds.

In the second stage (Fig.~\ref{fig_framework}(d), Step~1), the best-performing model $S^{(r)}_{\theta^*}$ selected by validation performs sliding-window inference on all unlabeled volumes to generate new pseudo-labels $\hat{y}^{(r)}_{u,\text{raw}}$ along with per-sample uncertainty estimates. We compute the average voxel-wise entropy of the softmax prediction as the sample-level uncertainty:
\begin{equation}
\mathcal{U}^{(r)}_u = \frac{-1}{|\Omega|} \sum_{v \in \Omega} \sum_{c=0}^{C_{\text{cls}}-1} P^{(r)}_u[c, v] \log P^{(r)}_u[c, v]
\label{eq:uncertainty}
\end{equation}
where $\Omega$ denotes the set of all voxels and $P^{(r)}_u = \text{softmax}(S^{(r)}_{\theta^*}(x_u))$ is the model's probabilistic output.

In the third stage (Fig.~\ref{fig_framework}(d), Step~2), the raw pseudo-labels are refined by leveraging the inter-sample relationships captured by the foundation model's global features. Using a pre-defined uncertainty quantile $q_{\text{unc}}$, the unlabeled set is partitioned into a \textit{certain set} $\mathcal{D}^{\text{cert}}_{U}$ (samples with uncertainty below the quantile threshold, plus the labeled sample) and an \textit{uncertain set} $\mathcal{D}^{\text{unc}}_{U}$ (the remaining high-uncertainty samples). For each uncertain sample $x_u \in \mathcal{D}^{\text{unc}}_{U}$, we query the $K$ nearest neighbors from $\mathcal{D}^{\text{cert}}_{U}$ using cosine similarity in the foundation model's global feature space (Eq.~\ref{eq:global_feat}). Let $\{x_{n_1}, \ldots, x_{n_K}\}$ denote the $K$ nearest certain neighbors with similarity weights $w_j = \max(0, \mathbf{f}^{g}_u \cdot \mathbf{f}^{g}_{n_j})$. The refined pseudo-label is obtained via weighted voting:
\begin{equation}
\hat{y}^{(r)}_{u,\text{refined}}[v] = \arg\max_{c} \frac{\sum_{j=1}^{K} w_j \cdot \mathbb{I}(\hat{y}^{(r)}_{n_j,\text{raw}}[v] = c)}{\sum_{j=1}^{K} w_j + \epsilon}
\label{eq:knn_refine}
\end{equation}
For certain samples, the raw pseudo-labels are retained unchanged. The refined pseudo-labels $\hat{\mathcal{Y}}^{(r)}$ are then used as input for round $r+1$.

This iterative process creates a virtuous cycle: improved pseudo-labels lead to better-trained models, which in turn generate more accurate pseudo-labels. Importantly, the foundation model serves solely as an offline feature extractor for initial pseudo-label propagation, without any involvement during the specialist model training. This fully decoupled design eliminates the computational overhead of online generalist inference in SemiSAM+ while maintaining the benefit of leveraging the foundation model's pre-trained representations for quality control.

\begin{table}[t]
	\caption{Comparison with state-of-the-art segmentation methods on LA segmentation under the one-label setting. Best results are in \textbf{bold}. SemiSAM-Based approaches are all implemented in ( MT / SAM-Med3D ) setting. } \label{Table_sota_la}
	\centering
	\footnotesize
	\renewcommand\arraystretch{1.5}
	\begin{tabular}{c|c|c}
		\hline \hline
		Model Type & Method & Dice$\uparrow$[\%] \\ \hline
		FS Baseline & 3D U-Net~\citep{cciccek20163d} & 37.97$\pm$21.80 \\ \hline
		\multirow{5}{*}{SAM Adaptions} & AutoSAM~\citep{autosam} & 57.99$\pm$13.72 \\
		& SAMed~\citep{SAMed} & 64.16$\pm$10.77 \\
		& MA-SAM~\citep{chen2024ma} & 36.67$\pm$10.68 \\
		& H-SAM~\citep{cheng2024unleashing} & 67.73$\pm$7.47 \\
		& Auto-SAM-Med3D & 61.07$\pm$12.30 \\ \hline
		\multirow{4}{*}{SSL Methods} & MT~\citep{tarvainen2017mean} & 
        40.64$\pm$23.82 \\
        & UGMCL~\citep{zhang2023uncertainty} & 44.08$\pm$23.17 \\
		& ACMT~\citep{xu2023ambiguity} & 44.17$\pm$21.82 \\
		& SDCL$^\dagger$~\citep{song2024sdcl} & 59.75$\pm$13.37 \\ \hline
		\multirow{3}{*}{SAM-Based SSL} & SemiSAM+ \citep{SemiSAM+}  & 48.91$\pm$22.79 \\
        & FM-MVSSL \citep{jiang2025foundation} & 51.52$\pm$22.08 \\ 
        & MVC-SemiSAM \citep{li2025foundation} & 50.74$\pm$22.52 \\ \hline
		\multirow{3}{*}{SemiSAM-O1} & MT / SAM-Med3D (R1) & 72.02$\pm$10.42 \\
		& MT / SAM-Med3D (R2) & \textbf{76.78$\pm$8.02} \\
		& MT / SAM-Med3D (R3) & 75.66$\pm$9.56 \\ \hline
		FS Upperbound & 3D U-Net~\citep{cciccek20163d} & 88.18$\pm$5.30 \\ \hline \hline
        \multicolumn{3}{l}{\footnotesize $^\dagger$ SDCL uses multi-network backbones different from other SSL methods.} 
	\end{tabular}
\end{table}

\begin{figure*}[t]
\centering
\includegraphics[width=\linewidth]{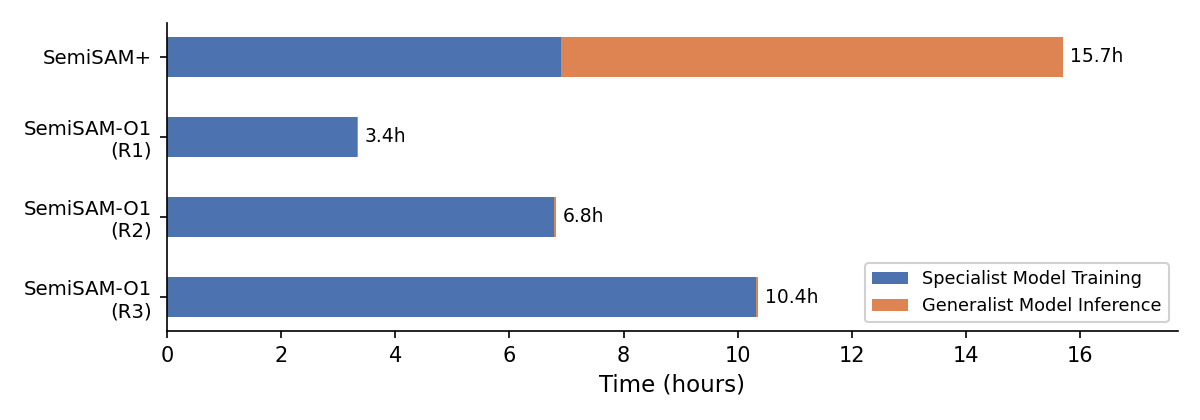}
\caption{Training time breakdown of SemiSAM+ and SemiSAM-O1 (R1-R3) on the LA dataset. Blue and orange segments indicate the time spent on specialist model training and generalist model inference, respectively.}
\label{fig_time}
\end{figure*}

\section{Experiments}

\subsection{Datasets}

To comprehensively evaluate SemiSAM-O1, we conduct experiments on two public benchmark datasets and two in-house clinical datasets, covering diverse imaging modalities and anatomical targets. All tasks are formulated as 3D segmentation problems under the one-label setting ($M=1$), where only a single annotated volume is available for supervision.

The first public dataset is the \textbf{Left Atrium (LA)} Segmentation Challenge dataset~\citep{xiong2021global}, which contains 100 3D gadolinium-enhanced MR imaging scans (GE-MRIs) with an isotropic resolution of $0.625 \times 0.625 \times 0.625\,\text{mm}^{3}$ and corresponding LA segmentation masks. We split the dataset into 75 scans for training, 5 for validation, and 20 for testing.
The second public dataset is the \textbf{BraTS 2019} dataset~\citep{hdtd-5j88-20} for brain tumor segmentation. Following~\citep{luo2022semi}, we investigate semi-supervised segmentation of whole tumors from FLAIR MRI images, as this modality can well characterize malignant tumors~\citep{zeineldin2020deepseg}. We use 250 scans for training, 25 for validation, and 60 for testing.

In addition to public benchmark datasets, we further evaluate our method on two in-house datasets. Notably, all in-house data correspond to unseen modalities or clinical tasks during the training of the adopted generalist model, which further verifies the generalization performance on new segmentation tasks in real-world scenarios. The data usage were approved by institutional ethical review boards.
The first in-house dataset \textbf{PETS} contains 100 FDG-PET images with expert-examined annotations of the heart for quantitative evaluation of cardiac metabolism. The PET images were reconstructed with $4.07 \times 4.07 \times 3\,\text{mm}^{3}$ voxels. We use 40 scans for training, 10 for validation, and 50 for testing.
Notably, SAM-Med3D was exclusively pre-trained on anatomical structural modalities and has never encountered functional PET data.
The second in-house dataset is \textbf{RT-EC} with manually delineated clinical target volume (CTV) and planning target volume (PTV) by experienced radiation oncologists.
RT-EC consists of 84 CT scans from endometrial cancer patients, which is divided into 50 training, 10 validation, and 24 testing cases. 
All volumes are resampled to a uniform spacing of 
$1.0 \times 1.0 \times 3.0 \text{mm}^{3}$ and intensity-normalized to zero mean and unit variance. CTV and PTV typically occupies a small fraction of the pelvic region, making these datasets particularly challenging for semi-supervised methods due to the severe class imbalance and the complex, irregular geometry of radiotherapy target volumes.

\begin{figure}[t]
\centering
\includegraphics[width=\linewidth]{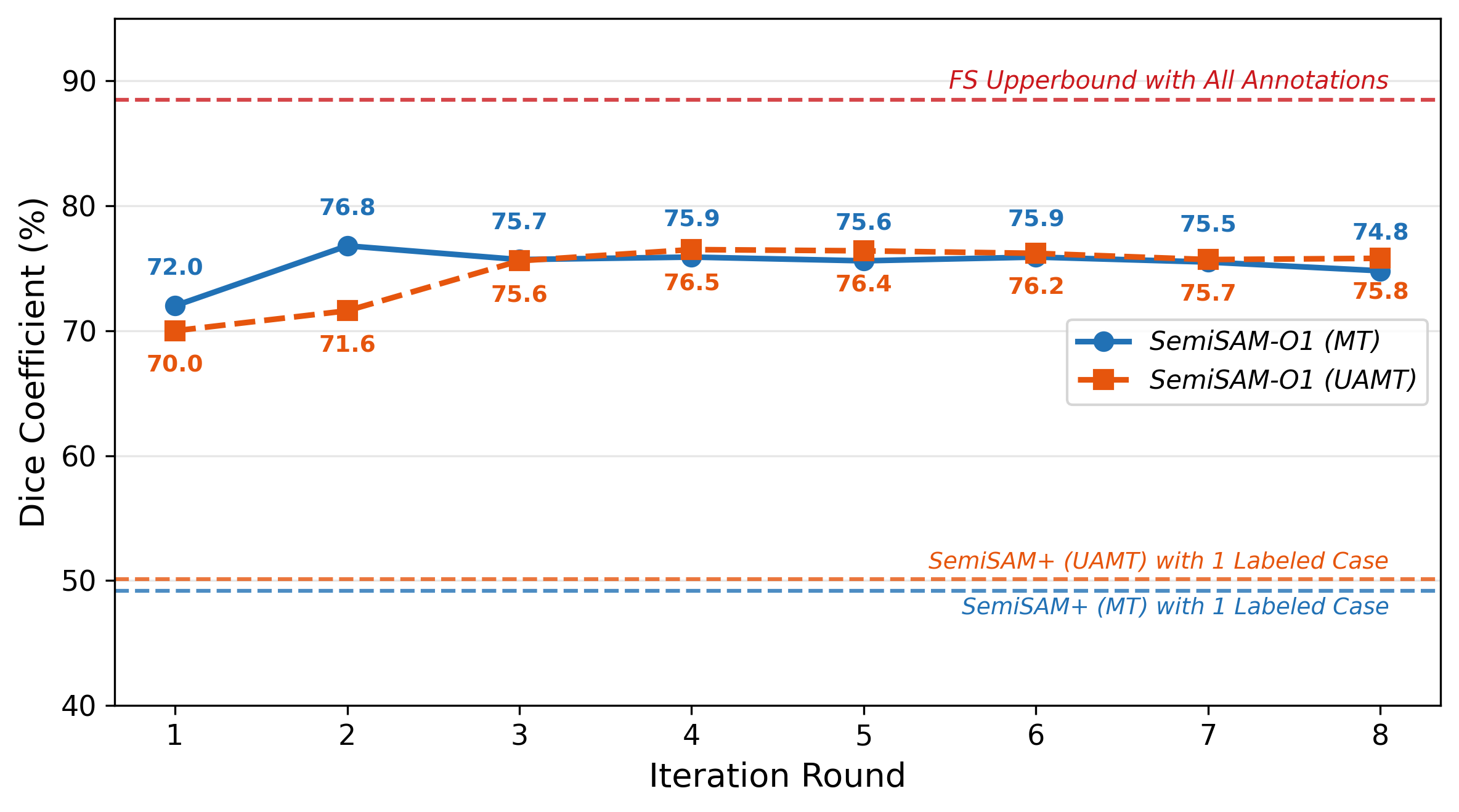}
\caption{Effect of the number of iterative refinement rounds of SemiSAM-O1 on segmentation performance (Dice) using the MT and UA-MT backbones on the LA dataset.}
\label{fig_round_ablation}
\end{figure}

\begin{figure}[t]
\centering
\includegraphics[width=\linewidth]{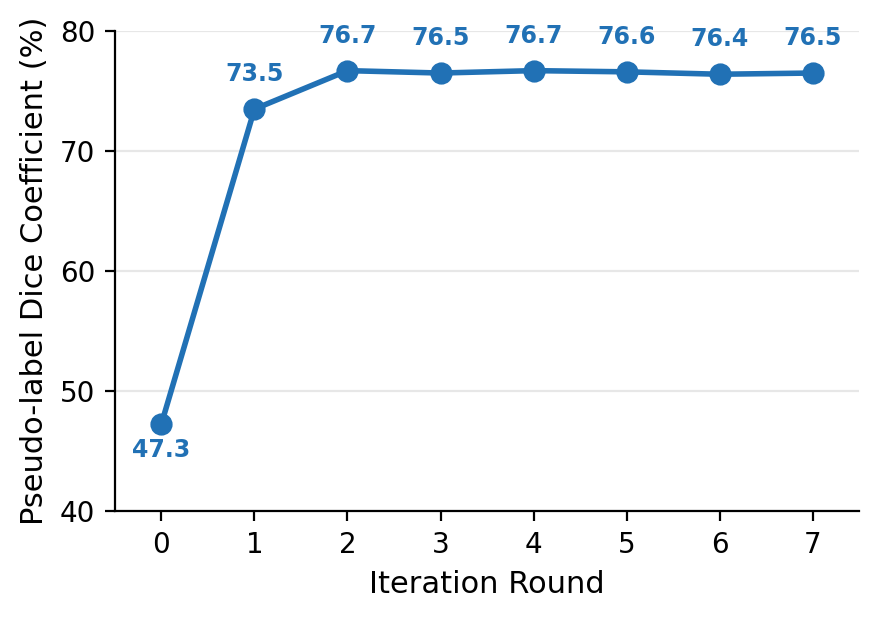}
\caption{Pseudo-label quality measured by Dice against ground truth annotation on unlabeled training samples using the MT backbone on the LA dataset. The quality rises sharply from R0 to R2 and then saturates, indicating that the refinement reaches a ceiling after the first two rounds.}
\label{fig_pl_quality}
\end{figure}

\begin{figure*}[t]
\centering
\includegraphics[width=\linewidth]{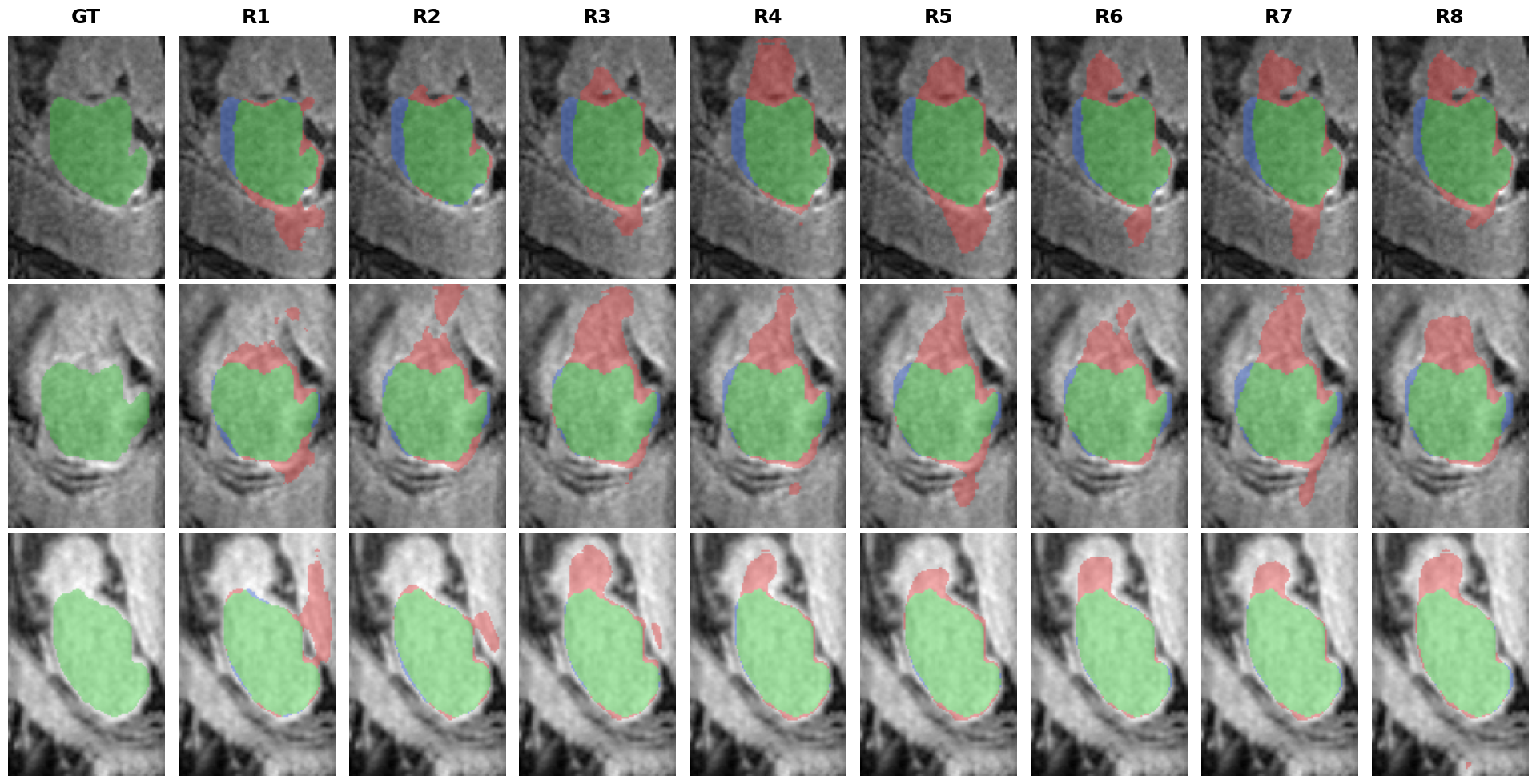}
\caption{Qualitative visualization of segmentation predictions from R1 to R8 on representative test cases. The green area, red area, and blue area represent True Positive (TP), False Positive (FP), and False Negative (FN), respectively.}
\label{fig_round_vis}
\end{figure*}

\subsection{Implementation Details}
All experiments are implemented in PyTorch on an NVIDIA A100 80GB GPU.
The specialist segmentation model backbone is 3D U-Net~\citep{cciccek20163d}, and the generalist foundation model is the latest version SAM-Med3D-turbo~\citep{SAM-Med3D}.
To demonstrate the generality of SemiSAM-O1, we integrate it with four representative SSL backbones: Mean Teacher (MT)~\citep{tarvainen2017mean}, Uncertainty-Aware Mean Teacher (UA-MT)~\citep{yu2019uncertainty}, Deep Adversarial Network (DAN)~\citep{zhang2017deep}, and Dual-Task Consistency (DTC)~\citep{luo2021semi}. Note that DTC employs a dual-head 3D U-Net with an additional signed distance function (SDF) branch.
We use the SGD optimizer with an initial learning rate of 0.01, momentum of 0.9, and weight decay of $10^{-4}$.
The learning rate follows a polynomial decay schedule: $\text{lr} = \text{lr}_0 \times (1 - t/t_{\max})^{0.9}$.
The batch size is set to 2 (1 labeled + 1 unlabeled).
Input patches of size $128 \times 128 \times 128$ are randomly cropped, with random flipping and rotation for data augmentation.
Each iterative round consists of 15,000 training iterations.
The pseudo-label weight $\alpha^{(r)}$ linearly ramps up from 0 to 1 during the first 30\% of iterations per round.
For KNN refinement, we set $K=5$ and the uncertainty quantile threshold $q_{\text{unc}}=0.9$.
During inference, segmentation results are obtained using a sliding-window strategy with a stride of 64 voxels.
We report four standard metrics: Dice similarity coefficient (Dice), Jaccard index (Jaccard), 95\% Hausdorff Distance (95HD), and Average Surface Distance (ASD).

\subsection{Adaptability across Different SSL Backbones}

Table~\ref{Table_oneseg} presents the results of integrating SemiSAM-O1 with four SSL representative backbones on the LA dataset under the one-label setting. Taking UA-MT as an example, the SSL baseline achieves only 41.73\% Dice, and incorporating SemiSAM+ improves it to 50.10\%. By contrast, SemiSAM-O1 substantially boosts the performance through iterative refinement. The Dice score increases from 70.03\% (R1) to 71.64\% (R2) and further to 75.60\% (R3), representing a 25.50\% absolute Dice improvement over SemiSAM+. Notably, the standard deviation also decreases from $\pm$22.83 to $\pm$8.26, indicating that the iterative pipeline produces more stable predictions across test cases. The 95HD metric improves from 37.02 to 24.75, confirming that the boundary quality is also substantially refined. Similar improvements are observed for other SSL backbones. The consistent improvement from R1 to R3 validates the effectiveness of the iterative pseudo-label refinement strategy: each round of model training, pseudo-label update, and KNN correction forms a virtuous cycle that progressively enhances segmentation quality.

\subsection{Comparison with State-of-the-art Methods}
Table~\ref{Table_sota_la} compares SemiSAM-O1 with state-of-the-art semi-supervised learning (SSL) methods and SAM-based fine-tuning approaches on the LA dataset under the one-label setting.
Among SAM fine-tuning methods, H-SAM \citep{cheng2024unleashing} achieves the highest Dice score of 67.73\%, benefiting from hierarchical prompt adaptation. However, these fine-tuning approaches still rely on the generalist SAM model during inference, leading to substantial computational overhead.
Among conventional SSL methods, SDCL \citep{song2024sdcl} achieves the best performance of 59.75\% by leveraging multiple network backbones for cross-model consistency. SemiSAM+ improves the performance of vanilla naive MT framework to 48.91\% by introducing foundation model-driven consistency regularization. We further reproduce two recent SSL frameworks \citep{jiang2025foundation, li2025foundation} adapted from SemiSAM+ for comparison. Although these methods achieve improvements over SemiSAM+, they still remain far from the fully supervised upper bound.
Building upon the same MT backbone, SemiSAM-O1 achieves 76.78\% Dice at R2, surpassing all competing methods. This result demonstrates that the proposed iterative pseudo-label refinement pipeline can effectively unlock the representation capability of the foundation model beyond its prompting interface, substantially narrowing the gap toward fully supervised performance. Meanwhile, during inference, our method only requires a specialist model 3D U-Net, avoiding the heavy computational cost associated with SAM-based methods.

\subsection{Training Efficiency Analysis}

A practical concern of the iterative pipeline of SemiSAM+ is its computational cost. Fig.~\ref{fig_time} presents the training time breakdown of SemiSAM+ and SemiSAM-O1 (R1-R3) on the LA dataset using the UA-MT backbone, where time is decomposed into specialist model training and generalist model inference.
In SemiSAM+, the generalist foundation model performs online inference throughout the entire training process, contributing 8.8 hours out of the total 15.7 hours (56\%). This heavy generalist overhead stems from the need to repeatedly invoke the 3D foundation model at full resolution during training to compute the consistency regularization signal. This issue is further exacerbated in multi-class segmentation tasks, as the generalist model is required to conduct separate inference for each individual category.

By contrast, SemiSAM-O1 decouples the generalist model from the training loop. The foundation model is only invoked offline at three stages: (1) once before training for feature extraction and prototype-based pseudo-label initialization ($\sim$1 minute), and (2) briefly between rounds for sliding-window pseudo-label generation and KNN refinement ($\sim$20 seconds per round). As a result, the generalist inference time is negligible ($<$0.1 hours across all three rounds), and the total training time is almost entirely determined by the specialist model training.
Notably, even after running three full rounds of iterative refinement (45{,}000 cumulative training iterations), SemiSAM-O1 (R3) requires only 10.4 hours, which is 34\% faster than SemiSAM+ (15.7 hours with 30{,}000 iterations). This efficiency gain is achieved while simultaneously delivering substantially higher segmentation performance (75.60\% vs.\ 50.10\% Dice). The results demonstrate that SemiSAM-O1 not only improves accuracy but also enhances training efficiency by shifting the generalist model's role from expensive online inference to lightweight offline feature extraction.

\subsection{Effect of Iterative Refinement Rounds}

A natural question arising from the iterative pipeline is: \textit{does performance continue to improve with more refinement rounds?} To investigate this, we extend the number of rounds from $R=3$ to $R=8$ using both the MT and UAMT backbones on the LA dataset. As shown in Fig.~\ref{fig_round_ablation}, both backbones exhibit a consistent pattern: rapid improvement in the early rounds followed by saturation at a stable plateau. The two backbones converge to this plateau at different speeds owing to their inherent characteristics, with MT reaching it by R2 and UAMT by R4, but ultimately arrive at comparable performance levels. Despite the saturation, all rounds remain substantially above the respective SemiSAM+ (48.91\% for MT, 50.10\% for UAMT), confirming that the iterative pipeline consistently and significantly outperforms conventional approaches regardless of the SSL backbone or the number of rounds.

We attribute this pattern to confirmation bias in the iterative pipeline. In early rounds, the pseudo-labels generated by prototype propagation contain many obvious errors, and the KNN refinement step can effectively correct them, leading to rapid performance gains. However, in later rounds, the remaining errors in the pseudo-labels tend to be systematic. Although the model is re-initialized at each round, the bias is transmitted through the pseudo-labels themselves.

\begin{figure}[t]
  \centering
  \includegraphics[width=\linewidth]{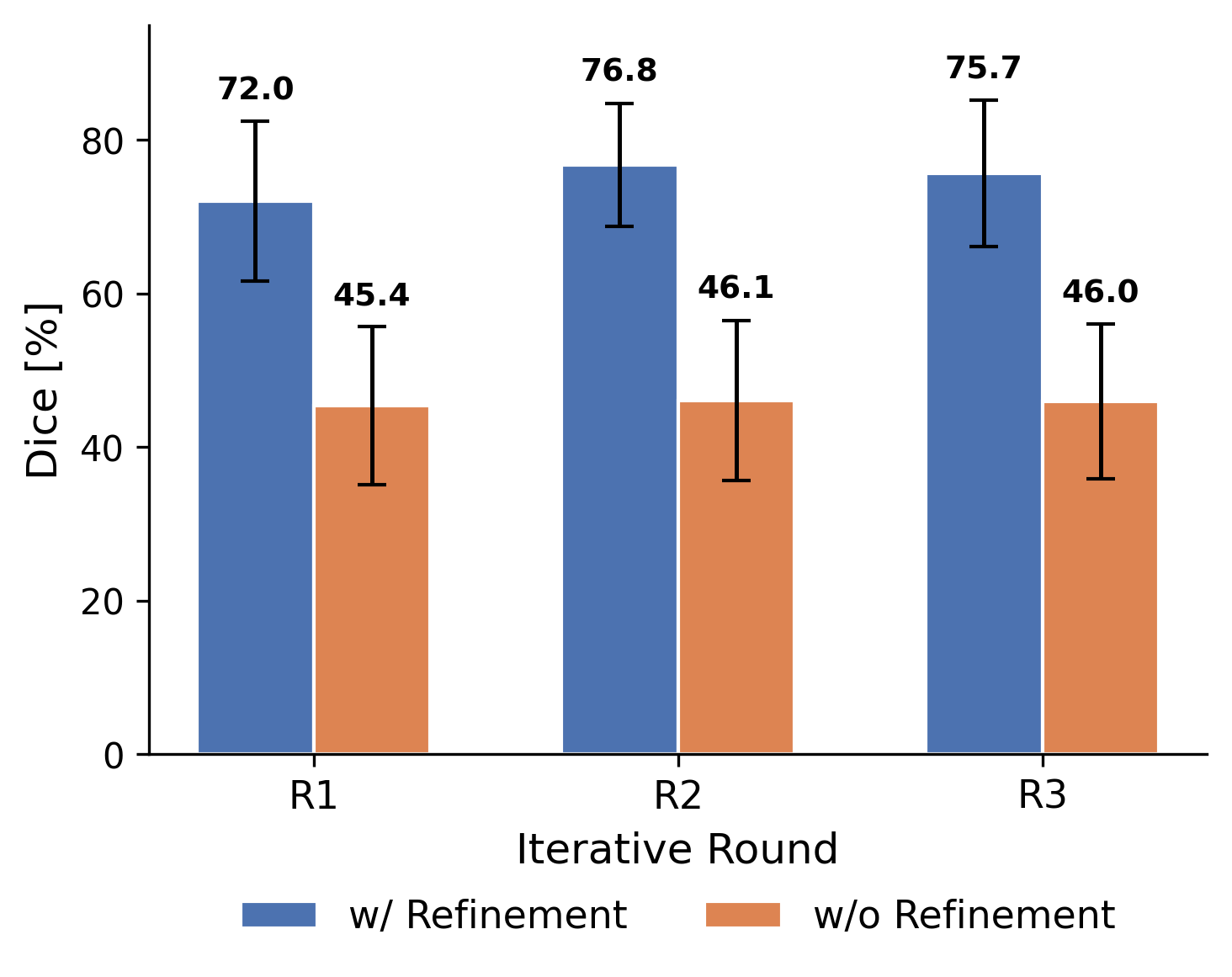}
  \caption{Ablation study on the pseudo-label refinement step. Removing the uncertainty-guided KNN refinement leads to performance degradation and eliminates the iterative improvement across rounds.} 
  \label{fig_knn_ablation}
\end{figure}

\begin{figure}[t]
  \centering
  \includegraphics[width=\linewidth]{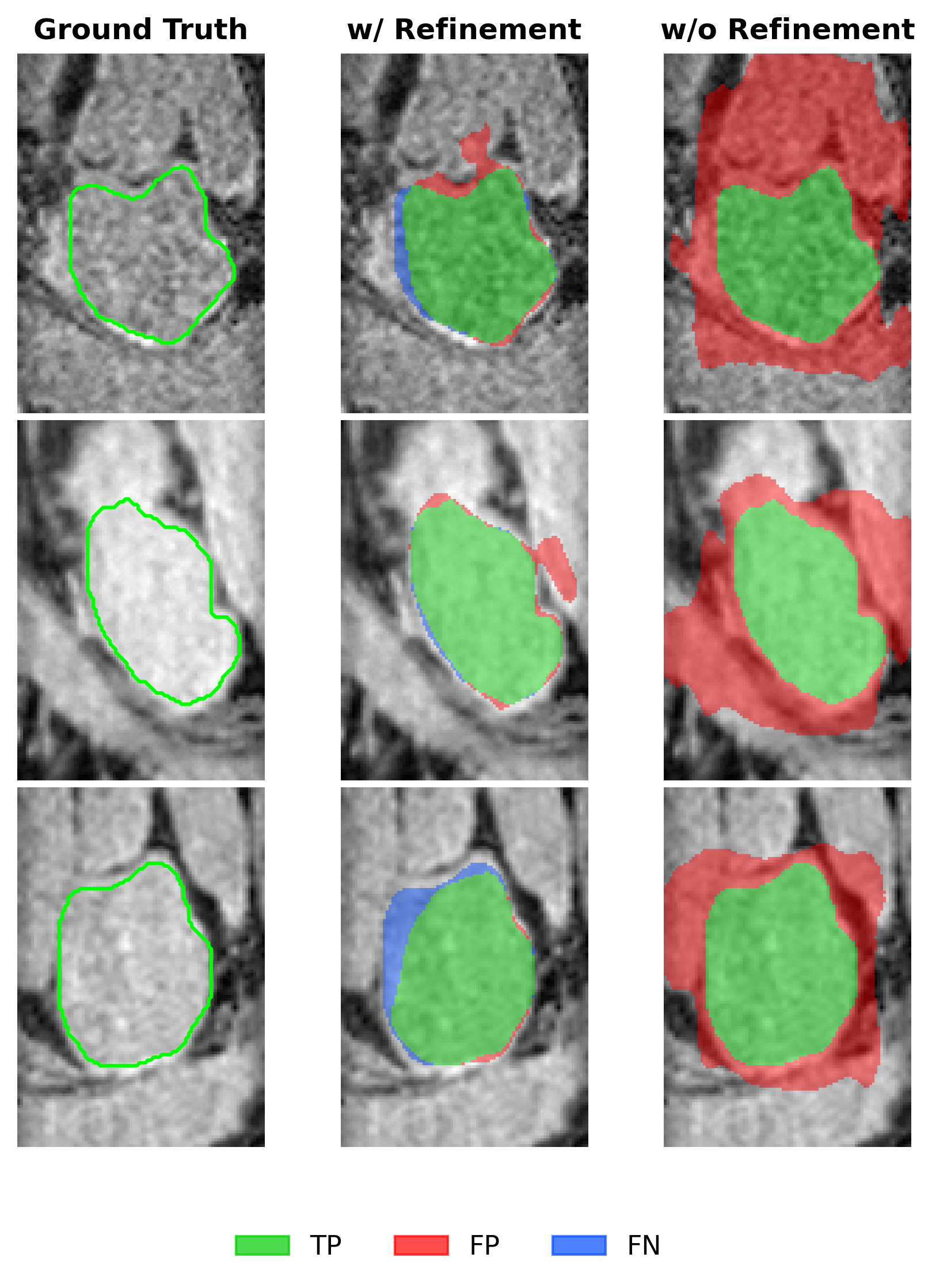}
  \caption{Qualitative comparison of segmentation results with and without the pseudo-label refinement step.} 
  \label{fig_knn_vis}
\end{figure}

To further verify this hypothesis, we provide two additional pieces of evidence on MT backbone. First, Fig.~\ref{fig_pl_quality} tracks the pseudo-label quality across rounds by computing the Dice coefficient between the generated pseudo-labels and ground truth on the unlabeled training samples. The pseudo-label Dice rises sharply from 47.3\% at R0 (prototype propagation) to 73.5\% after R1, and further to 76.7\% after R2, before saturating around 76.4--76.7\% for the remaining rounds. This plateau closely mirrors the test Dice saturation observed in Fig.~\ref{fig_round_ablation}, directly confirming that the performance ceiling in later rounds originates from the pseudo-label quality reaching its limit, rather than from the model's capacity or training instability.
Second, Fig.~\ref{fig_round_vis} visualizes the segmentation predictions from R1 to R8 on three test cases. At R1, the predictions contain substantial false-positive and false-negative regions. After the first round of pseudo-label refinement, R2 shows markedly improved boundary quality with reduced errors across all cases. However, starting from R3, segmentation errors in specific regions become locked in. This spatial consistency of errors across rounds provides direct visual evidence of confirmation bias.
These findings demonstrate that for the one-label setting, a small number of rounds ($R=2$ or $R=3$) offers the best trade-off between performance and computational cost.

\subsection{Ablation Analysis of Pseudo-Label Refinement}

The uncertainty-guided KNN refinement step is a core component of SemiSAM-O1 that corrects high-uncertainty pseudo-labels between rounds by leveraging inter-sample relationships in the foundation model's feature space. To quantify its contribution, we conduct an ablation study by removing the KNN refinement step while keeping all other components unchanged. All experiments use the MT backbone on the LA dataset.  As shown in Fig.~\ref{fig_knn_ablation}, removing the refinement step leads to a dramatic performance drop across all rounds. With refinement, the Dice score increases from 72.0\% (R1) to 76.8\% (R2), reflecting the progressive improvement enabled by the iterative pipeline. Without  refinement, the Dice remains stagnant around 45-46\%. This result reveals that the KNN refinement is essential for the iterative pipeline to function effectively. Without it, the pseudo-labels generated by the model alone are insufficiently accurate to drive meaningful improvement across rounds, and the  iterative process degenerates into repeated training on similarly noisy supervision. Second, the refinement step serves as the primary mechanism  that breaks confirmation bias in the early rounds, enabling the virtuous cycle of improving pseudo-labels and model performance. Fig.~\ref{fig_knn_vis} provides qualitative evidence that further corroborates these findings, confirming that the unrefined pseudo-labels propagate systematic errors that the model cannot self-correct through training alone.

We further investigate the sensitivity of the two key hyperparameters in the refinement module, the number of neighbors $K$ (Table~\ref{Table_knn_ablation_K}) and the uncertainty threshold $q_{\text{unc}}$ (Table~\ref{Table_knn_ablation_qunc}). Overall, SemiSAM-O1 demonstrates strong robustness to both hyperparameters, with consistently high performance across a broad range of settings. For the neighborhood size, using a very small $K$ makes the refinement more susceptible to noisy neighbors, whereas an excessively large $K$ introduces less relevant samples into the voting process, slightly degrading the pseudo-label quality. Moderate neighborhood sizes consistently achieve the best performance. Regarding the uncertainty threshold, small values classify too many samples as uncertain, causing reliable predictions to be unnecessarily replaced by neighbor voting. Conversely, setting $q_{\text{unc}}=1.0$ removes uncertainty filtering entirely, allowing unreliable pseudo-labels to propagate into subsequent rounds. Selecting only the top 10\% most uncertain samples for refinement provides the best balance between preserving confident predictions and correcting unreliable ones, leading to the optimal overall performance.

\begin{table}[t]
\centering
\footnotesize
\renewcommand\arraystretch{1.3}
\caption{Ablation study on hyperparameter $K$ of the uncertainty-guided KNN refinement step on the LA dataset using the MT backbone.}
\label{Table_knn_ablation_K}
\begin{tabular}{c|c|cccc}
\hline \hline
$K$          & Round & Dice$\uparrow$[\%]       & Jaccard$\uparrow$[\%]    & 95HD$\downarrow$     & ASD$\downarrow$      \\ \hline
\multirow{3}{*}{1}   & R1    & 71.57$\pm$10.96          & 56.82$\pm$12.89          & 26.64$\pm$11.91      & 8.60$\pm$4.24        \\
             & R2    & 76.06$\pm$8.85           & 62.14$\pm$10.94          & 20.29$\pm$10.76      & 6.47$\pm$3.88        \\
             & R3    & 75.60$\pm$9.40           & 61.64$\pm$11.60          & 23.22$\pm$11.25      & 6.96$\pm$4.25        \\ \hline
\multirow{3}{*}{3}   & R1    & 71.15$\pm$10.99          & 56.32$\pm$12.85          & 27.89$\pm$13.35      & 8.92$\pm$4.39        \\
             & R2    & 75.95$\pm$9.15           & 62.05$\pm$11.34          & 21.74$\pm$11.86      & 6.76$\pm$4.46        \\
             & R3    & 76.50$\pm$9.25           & 62.81$\pm$11.65          & 20.62$\pm$11.63      & 6.16$\pm$4.51        \\ \hline
\multirow{3}{*}{\textbf{5}} & R1 & 72.02$\pm$10.42 & 57.27$\pm$12.28 & 28.25$\pm$13.43 & 8.87$\pm$4.45 \\
             & R2    & \textbf{76.78$\pm$8.02}  & \textbf{62.96$\pm$9.96}  & \textbf{20.24$\pm$9.83}  & \textbf{5.84$\pm$3.00}  \\
             & R3    & 75.66$\pm$9.56           & 61.73$\pm$11.64          & 21.21$\pm$11.12      & 6.63$\pm$4.18        \\ \hline
\multirow{3}{*}{7}   & R1    & 71.59$\pm$11.17          & 56.88$\pm$13.01          & 24.66$\pm$11.64      & 8.02$\pm$3.99        \\
             & R2    & 76.46$\pm$9.13           & 62.72$\pm$11.36          & 19.50$\pm$11.81      & 6.10$\pm$4.45        \\
             & R3    & 76.15$\pm$9.47           & 62.39$\pm$11.80          & 20.91$\pm$11.63      & 6.20$\pm$4.40        \\ \hline
\multirow{3}{*}{10}  & R1    & 70.55$\pm$11.49          & 55.67$\pm$13.30          & 30.19$\pm$14.17      & 9.31$\pm$4.92        \\
             & R2    & 75.94$\pm$9.63           & 62.13$\pm$11.81          & 21.09$\pm$12.62      & 6.50$\pm$4.75        \\
             & R3    & 75.73$\pm$9.39           & 61.83$\pm$11.75          & 20.01$\pm$11.05      & 5.87$\pm$4.25        \\
\hline \hline
\end{tabular}
\end{table}

\begin{table}[t]
\centering
\footnotesize
\renewcommand\arraystretch{1.3}
\caption{Ablation study on hyperparameter $q_\text{unc}$ of the uncertainty-guided KNN refinement step on the LA dataset using the MT backbone.}
\setlength{\tabcolsep}{5pt} 
\label{Table_knn_ablation_qunc}
\begin{tabular}{c|c|cccc}
\hline \hline
$q_\text{unc}$ & Round & Dice$\uparrow$[\%]       & Jaccard$\uparrow$[\%]    & 95HD$\downarrow$     & ASD$\downarrow$      \\ \hline
0.5           & R1    & 69.70$\pm$11.30          & 54.61$\pm$12.93          & 32.02$\pm$14.88          & 9.84$\pm$4.68          \\
              & R2    & 74.57$\pm$10.36          & 60.48$\pm$12.39          & 20.29$\pm$11.81          & 6.22$\pm$4.80          \\
              & R3    & 74.59$\pm$10.32          & 60.52$\pm$12.67          & 21.36$\pm$10.46          & 6.29$\pm$4.36          \\ \hline
0.7           & R1    & 69.39$\pm$11.46          & 54.28$\pm$13.08          & 29.62$\pm$12.91          & 9.64$\pm$4.63          \\
              & R2    & 74.57$\pm$11.26          & 60.66$\pm$13.46          & 21.43$\pm$11.90          & 6.68$\pm$4.83          \\
              & R3    & 74.19$\pm$11.34          & 60.18$\pm$13.41          & 21.97$\pm$11.61          & 6.72$\pm$5.07          \\ \hline
\textbf{0.9}  & R1    & 72.02$\pm$10.42          & 57.27$\pm$12.28          & 28.25$\pm$13.43          & 8.87$\pm$4.45          \\
              & R2    & \textbf{76.78$\pm$8.02}  & \textbf{62.96$\pm$9.96}  & \textbf{20.24$\pm$9.83}  & \textbf{5.84$\pm$3.00}  \\
              & R3    & 75.66$\pm$9.56           & 61.73$\pm$11.64          & 21.21$\pm$11.12          & 6.63$\pm$4.18          \\ \hline
0.95          & R1    & 71.52$\pm$11.48          & 56.85$\pm$13.29          & 27.76$\pm$13.62          & 8.78$\pm$4.71          \\
              & R2    & 76.27$\pm$9.59           & 62.57$\pm$11.90          & 20.33$\pm$12.30          & 6.36$\pm$4.56          \\
              & R3    & 76.27$\pm$9.63           & 62.56$\pm$11.82          & 20.19$\pm$11.76          & 6.36$\pm$4.45          \\ \hline
1.0           & R1    & 70.47$\pm$11.12          & 55.52$\pm$12.96          & 31.40$\pm$15.37          & 9.57$\pm$5.06          \\
              & R2    & 72.57$\pm$9.49           & 57.81$\pm$11.51          & 23.27$\pm$11.65          & 7.42$\pm$4.14          \\
              & R3    & 75.14$\pm$9.95           & 61.15$\pm$12.23          & 20.83$\pm$10.80          & 6.40$\pm$4.06          \\
\hline \hline
\end{tabular}
\end{table}

\begin{table*}[t]
\caption{Robustness analysis with different labeled template selections on the LA dataset using the MT backbone. Three training cases with different anatomical characteristics are selected as the labeled template. In addition to the segmentation performance across refinement rounds, we report the atrial volume, foreground-background contrast, and shape compactness of each template to characterize its difficulty. }
\label{Table_template_selection}
\centering
\small
\renewcommand\arraystretch{1.3}
\begin{tabular}{c|c|c|c|c|cccc}
\hline\hline
\multirow{2}{*}{Template} &
\multirow{2}{*}{Volume} &
\multirow{2}{*}{Contrast} &
\multirow{2}{*}{Compactness} &
\multirow{2}{*}{Round} &
\multicolumn{4}{c}{Segmentation Performance} \\
\cline{6-9}
 & & & & &
Dice$\uparrow$[\%] &
Jaccard$\uparrow$[\%] &
95HD$\downarrow$ &
ASD$\downarrow$ \\
\hline
\multirow{3}{*}{Case 1}
& \multirow{3}{*}{154,296 (Small)}
& \multirow{3}{*}{2.42 (High)}
& \multirow{3}{*}{0.197 (Irregular)}
& R1 & 72.02$\pm$10.42 & 57.27$\pm$12.28 & 28.25$\pm$13.43 & 8.87$\pm$4.45 \\
&&&& R2 & \textbf{76.78$\pm$8.02} & \textbf{62.96$\pm$9.96} & \textbf{20.24$\pm$9.83} & \textbf{5.84$\pm$3.00} \\
&&&& R3 & 75.66$\pm$9.56 & 61.73$\pm$11.64 & 21.21$\pm$11.12 & 6.63$\pm$4.18 \\
\hline

\multirow{3}{*}{Case 2}
& \multirow{3}{*}{252,364 (Large)}
& \multirow{3}{*}{0.90 (Medium)}
& \multirow{3}{*}{0.283 (Compact)}
& R1 & 68.50$\pm$11.52 & 53.19$\pm$12.61 & 31.02$\pm$11.42 & 10.99$\pm$4.58 \\
&&&& R2 & 71.60$\pm$10.67 & 56.75$\pm$11.93 & 33.46$\pm$14.75 & 10.94$\pm$5.26 \\
&&&& R3 & \textbf{73.21$\pm$9.61} & \textbf{58.57$\pm$10.98} & \textbf{22.69$\pm$8.21} & \textbf{7.66$\pm$3.56} \\
\hline

\multirow{3}{*}{Case 3}
& \multirow{3}{*}{202,317 (Medium)}
& \multirow{3}{*}{0.75 (Low)}
& \multirow{3}{*}{0.198 (Irregular)}
& R1 & 65.52$\pm$9.55 & 49.46$\pm$10.41 & \textbf{30.14$\pm$8.97}  & 11.12$\pm$3.83 \\
&&&& R2 & \textbf{69.08$\pm$9.40} & \textbf{53.54$\pm$10.79} & 30.73$\pm$9.24  & \textbf{10.93$\pm$4.18} \\
&&&& R3 & 68.95$\pm$9.81 & 53.44$\pm$11.08 & 30.94$\pm$10.11 & 11.06$\pm$4.52 \\
\hline\hline
\end{tabular}
\end{table*}

\begin{table}[t]
\caption{Compatibility of SemiSAM-O1 with different medical foundation models on the LA dataset using the MT backbone. }
\label{tab:fm_generalization}
\centering
\footnotesize
\renewcommand{\arraystretch}{1.25}
\setlength{\tabcolsep}{3pt} 
\begin{tabular}{c|c|cccc}
\hline\hline
Backbone & Round &
Dice$\uparrow$[\%] &
Jaccard$\uparrow$[\%] &
95HD$\downarrow$ &
ASD$\downarrow$ \\
\hline
\multirow{3}{*}{SAM-Med3D}
& R1 & 72.02$\pm$10.42 & 57.27$\pm$12.28 & 28.25$\pm$13.43 & 8.87$\pm$4.45 \\
& R2 & \textbf{76.78$\pm$8.02} & \textbf{62.96$\pm$9.96} & \textbf{20.24$\pm$9.83} & \textbf{5.84$\pm$3.00} \\
& R3 & 75.66$\pm$9.56 & 61.73$\pm$11.64 & 21.21$\pm$11.12 & 6.63$\pm$4.18 \\
\hline
\multirow{3}{*}{nnInteractive}
& R1 & 81.13$\pm$7.20 & 68.82$\pm$9.39 & 19.35$\pm$13.33 & 4.55$\pm$5.03 \\
& R2 & \textbf{81.53$\pm$6.13} & \textbf{69.24$\pm$8.25} & \textbf{20.70$\pm$13.38} & \textbf{4.61$\pm$4.62} \\
& R3 & 80.43$\pm$7.33 & 67.87$\pm$9.81 & 22.10$\pm$13.70 & 5.00$\pm$5.10 \\
\hline
\multirow{3}{*}{SwinViT}
& R1 & 72.68$\pm$14.35 & 58.88$\pm$16.08 & 23.15$\pm$12.95 & 7.15$\pm$4.73 \\
& R2 & 77.06$\pm$9.88 & 63.69$\pm$12.58 & 19.64$\pm$11.18 & 5.59$\pm$4.19 \\
& R3 & \textbf{78.32$\pm$8.65} & \textbf{65.16$\pm$11.21} & \textbf{18.17$\pm$9.44} & \textbf{5.00$\pm$3.71} \\
\hline\hline
\end{tabular}
\end{table}

\subsection{Robustness to Different Template Selection}

To evaluate whether the proposed framework depends on the choice of the single labeled template, we randomly selected three different training cases as the only annotated sample for segmentation.
As summarized in Table~\ref{Table_template_selection}, the selected templates exhibit substantial diversity in anatomical characteristics. These differences cover representative variations commonly encountered in clinical practice.
Despite these pronounced differences, SemiSAM-O1 exhibits a consistent optimization pattern across different template selections. All three selections exhibit a consistent trend: the segmentation performance improves substantially after the first refinement round and reaches a stable plateau after R2-R3.

This demonstrates that SemiSAM-O1 is not overly dependent on a specific labeled template and can reliably propagate supervision from different reference cases. The remaining performance differences are expected because the anatomical completeness, image quality, and shape representativeness of the selected template directly affect the quality of the prototype initialization in Round 0. Nevertheless, the iterative pseudo-label refinement effectively mitigates these initial discrepancies, resulting in convergent performance across different template selections.

\begin{figure*}[t]
\centering
\includegraphics[width=\linewidth]{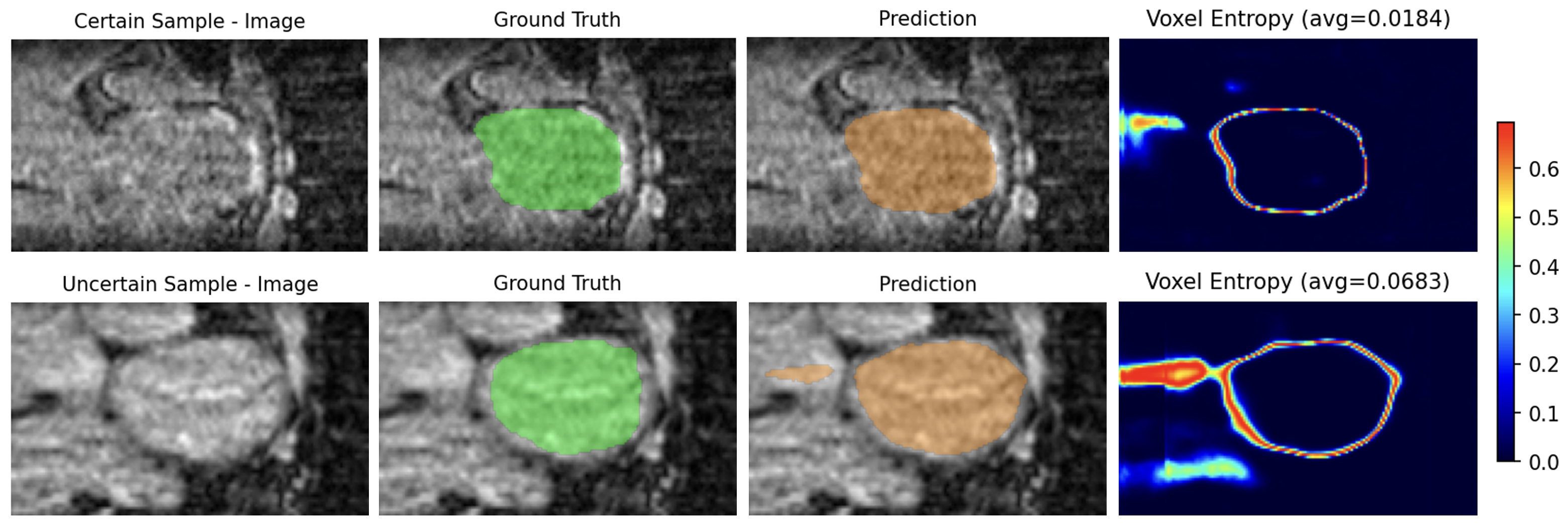}
\caption{Qualitative visualization of a certain sample and an uncertain sample. Each row presents the input image, ground truth annotation, model prediction, and voxel-wise entropy map.   High-entropy regions are confined to narrow segmentation boundaries in the certain sample, but extend across missegmented areas in the uncertain sample.}
\label{fig_uncvis}
\end{figure*}

\begin{figure}[t]
\centering
\includegraphics[width=\linewidth]{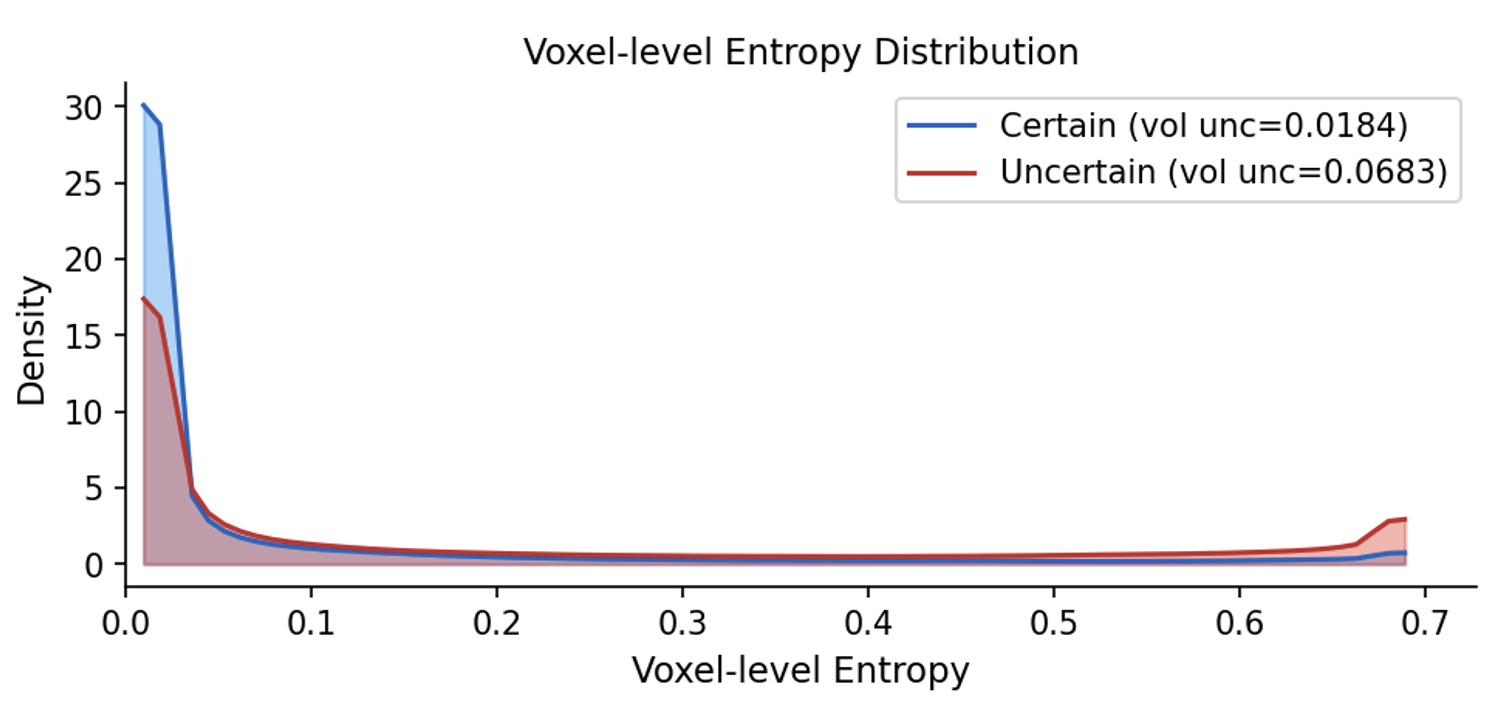}
\caption{Density distributions of a certain sample and an uncertain sample. The certain sample features a sharply peaked distribution concentrated at near-zero entropy, while the uncertain sample exhibits a markedly wider distribution with a prominent high-entropy tail.}
\label{fig_uncdis}
\end{figure}

\begin{figure}[t]
\centering
\includegraphics[width=\linewidth]{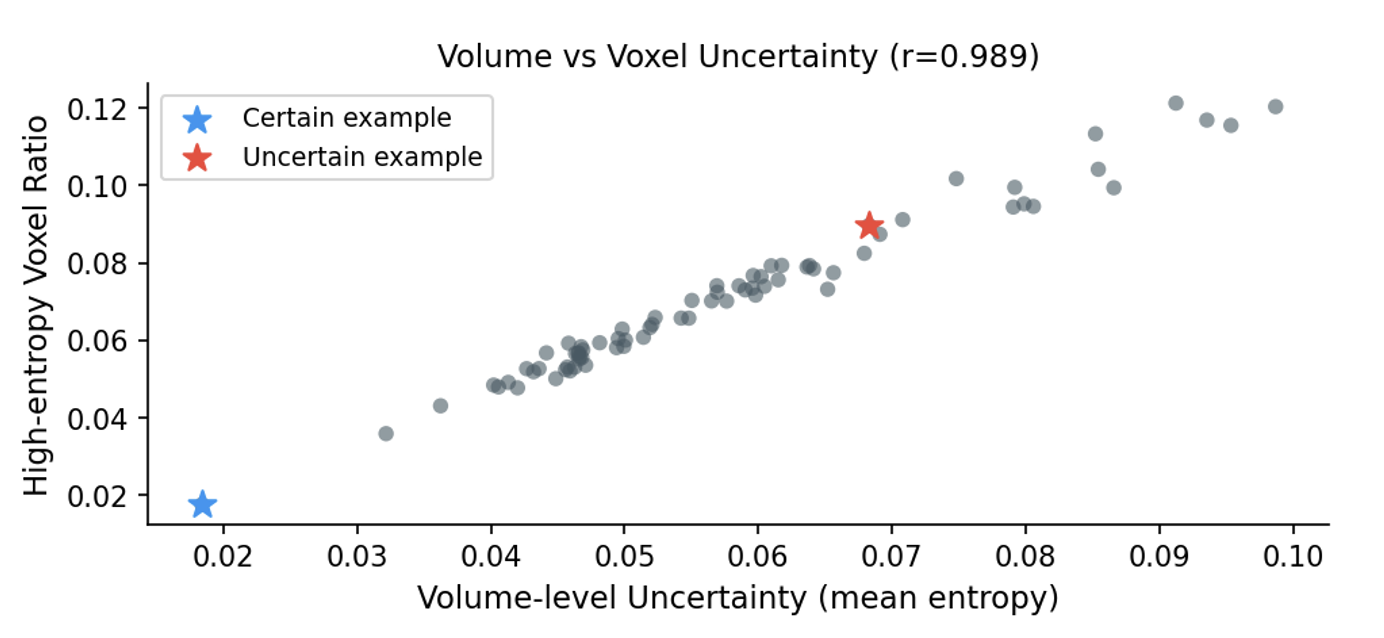}
\caption{Scatter plot of volume-level mean uncertainty versus the ratio of high-entropy voxels across all unlabeled training volumes.}
\label{fig_uncdis2}
\end{figure}

\subsection{Compatibility with Different Medical Foundation Models}

To further investigate whether SemiSAM-O1 is tied to a specific foundation model, we replace the default SAM-Med3D with two alternative medical foundation models, including promptable nnIteractive \citep{nninteractive} and non-promptable SwinViT \citep{tang2022self}. 
Table~\ref{tab:fm_generalization} summarizes the results. Despite differences in the feature representations produced by different foundation models, SemiSAM-O1 consistently achieves competitive segmentation performance across all settings. This observation suggests that the proposed framework does not rely on a particular foundation model, but instead benefits from the general semantic representations learned by large-scale medical foundation models.
Moreover, the iterative pseudo-label refinement consistently improves segmentation quality regardless of the selected feature extractor, indicating that the proposed refinement strategy is largely model-agnostic. Although stronger foundation models generally provide slightly better initialization due to more discriminative feature embeddings, the performance gap after iterative refinement remains relatively small. These results demonstrate that SemiSAM-O1 serves as a generic framework that can naturally accommodate future advances in medical foundation models without requiring modifications to the training pipeline.

\subsection{Voxel-Level Uncertainty Analysis}

The refinement procedure stratifies unlabeled volumes into certain and uncertain subsets based on sample-averaged prediction entropy. A core underlying assumption of this design is that volume-level uncertainty faithfully reflects per-voxel prediction reliability across the entire spatial extent.  Specifically, volumes classified as certain should deliver consistently confident predictions across most anatomical regions, rather than concealing localized high-uncertainty areas behind a low global average. We verify this assumption through a multi-scale uncertainty analysis on the left atrium dataset. 

Fig.~\ref{fig_uncvis} provides qualitative visualization of a representative certain volume and an uncertain volume, paired with their voxel-wise entropy maps. The certain volume yields an average voxel entropy of 0.0184, where elevated entropy values are strictly restricted to the narrow transition band along the segmentation boundary. The interior of the target structure exhibits uniformly low entropy, indicating stable and reliable predictions across the overwhelming majority of voxels. For the uncertain volume with an average voxel entropy of 0.0683, high-entropy regions extend far beyond the anatomical boundary and cover large erroneously segmented areas, including prominent false-positive regions adjacent to the left atrium. This qualitative pattern indicates that inferior prediction performance is not limited to isolated locations, but presents as a systemic uncertainty pattern distributed throughout the volume.
The density plot in Fig.~\ref{fig_uncdis} further quantifies the distributional discrepancy in voxel-level entropy between two representative volumes. The certain volume features a sharply peaked distribution concentrated at near-zero entropy, with only a negligible fraction of voxels falling into the high-entropy range. The uncertain volume exhibits a substantially wider distribution with a pronounced right tail, reflecting a substantial proportion of voxels with ambiguous class assignments. This distributional divergence confirms that sample-level mean entropy effectively captures the overall reliability of segmentation outputs.

To generalize this finding beyond individual cases, we examine the association between volume-level mean entropy and the proportion of high-entropy voxels across the entire unlabeled training set. As shown in Fig.~\ref{fig_uncdis2}, the two metrics exhibit an extremely strong linear association with a Pearson correlation coefficient of 0.989. Volumes with higher global uncertainty consistently contain a larger fraction of unreliable predictions, whereas volumes with low global uncertainty contain barely any high-uncertainty voxels.

In conclusion, these results validate the rationale of volume-level uncertainty partitioning in the refinement pipeline.  While boundary-adjacent uncertainty is ubiquitous even in well-performing segmentations, these regions account for a minimal proportion of total voxels and do not undermine the overall reliability of certain volumes as reference neighbors.  Volumes identified as uncertain at the sample level systematically suffer from widespread prediction degradation, making them appropriate candidates for KNN-based correction.  This analysis confirms that volume-level uncertainty serves as a robust proxy for per-voxel prediction quality, justifying its use for sample selection in the refinement stage.

\begin{figure}[t]
\centering
\includegraphics[width=\linewidth]{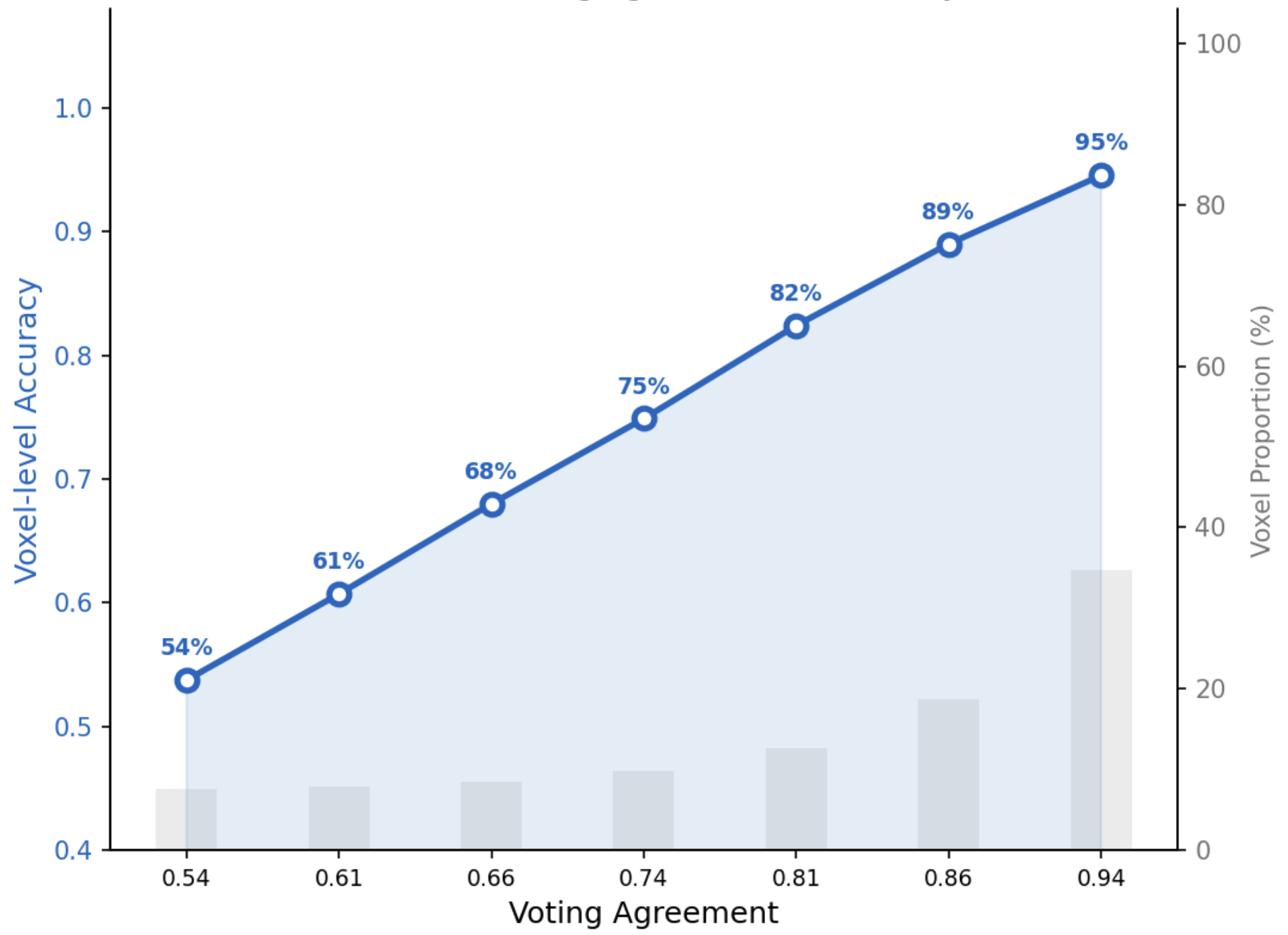}
\caption{Correlation between KNN voting agreement and voxel-level pseudo-label accuracy on the LA dataset. The blue curve denotes voxel-level classification accuracy against ground truth within each agreement bin, demonstrating a strong monotonic positive relationship. Gray bars represent the proportion of total voxels distributed across each agreement bin, showing that most voxels fall into high-agreement regions with reliable pseudo-label quality.}
\label{fig:knn_agreement_vs_accuracy}
\end{figure}

\subsection{Reliability Analysis of Voting Refinement }

A fundamental assumption underlying the uncertainty-guided KNN refinement is that volumes with high global feature similarity share consistent voxel-level anatomical correspondence, such that cross-sample label voting can reliably correct erroneous predictions. To empirically validate this assumption and quantify, we perform a voxel-level analysis to characterize the relationship between inter-neighbor voting agreement and prediction accuracy.
For each unlabeled training volume, we retrieve its nearest neighbors from the certain set based on global features extracted by the foundation model encoder, and conduct unweighted majority voting at every voxel position. Voting agreement is defined as the fraction of neighbors that vote for the majority class at a given voxel, which quantifies the level of consensus across structurally similar reference samples. All voxels in the unlabeled pool are discretized into bins according to their agreement values. For each bin, we compute two metrics: voxel-level classification accuracy measured against ground truth annotations, and the proportion of total voxels falling within that agreement range.

As shown in Fig.~\ref{fig:knn_agreement_vs_accuracy}, voting agreement exhibits a strong monotonic positive correlation with voxel-level accuracy.  Voxels with low agreement around 0.54 yield an accuracy of only 54\%, corresponding to highly ambiguous predictions where reference neighbors diverge substantially, approaching the performance of random assignment. By contrast, voxels with agreement above 0.94 achieve an accuracy of 95\%, confirming that consensus among anatomically similar neighbors serves as a reliable indicator of correct pseudo-label assignment. Across intermediate agreement levels, accuracy increases steadily following a smooth and predictable gradient.
The gray bars depict the distribution of voxel proportions across agreement bins. The vast majority of voxels concentrate in high-agreement regions, with the largest proportion residing in the highest agreement bin. This distribution indicates that KNN voting produces confident, consistent pseudo-labels for most spatial regions, while only a small fraction of voxels exhibit low inter-neighbor agreement. This characteristic aligns well with the design of our iterative refinement pipeline. For most voxels, cross-sample voting can directly generate high-quality labels, and only a small subset of low-agreement voxels requires further optimization across successive iterative rounds.

\begin{table*}[t]
	\caption{Cross-dataset extensive evaluation of SemiSAM-O1 under the one-label setting. All experiments use MT as the specialist model and SAM-Med3D as the generalist model. SemiSAM-O1 reports the best result across R1-R3.}
	\label{Table_cross}
	\centering
	\normalsize
	\renewcommand\arraystretch{1.3}
	\begin{tabular}{c|c|c|cccc}
		\hline \hline
		Dataset & Segmentation Target & Method & Dice$\uparrow$[\%] & Jaccard$\uparrow$[\%] & 95HD$\downarrow$[voxel] & ASD$\downarrow$[voxel] \\ \hline
		
		\multirow{6}{*}{BraTS 2019}
		& \multirow{6}{*}{Brain Tumor}
		& FS Baseline & 42.82$\pm$18.13 & 29.01$\pm$15.43 & 63.55$\pm$17.71 & 30.43$\pm$11.88 \\
		& & SSL Baseline & 52.56$\pm$15.25 & 37.13$\pm$14.51 & 58.80$\pm$21.11 & 25.41$\pm$13.78 \\
		& & SemiSAM+ & \textbf{68.53$\pm$22.22} & 56.22$\pm$24.39 & 38.81$\pm$29.05 & 16.66$\pm$15.93 \\
		& & SemiSAM-O1 (R1) & 61.60$\pm$25.62 & 49.37$\pm$26.52 & 41.46$\pm$29.04 & 18.42$\pm$15.96 \\
		& & SemiSAM-O1 (R2) & 65.85$\pm$23.98 & 53.60$\pm$25.50 & 37.32$\pm$29.02 & 16.38$\pm$15.72 \\
		& & SemiSAM-O1 (R3) & 68.50$\pm$23.48 & \textbf{56.51$\pm$25.05} & \textbf{33.97$\pm$27.67} & \textbf{14.26$\pm$14.80} \\ \hline
		
		\multirow{6}{*}{PETS}
		& \multirow{6}{*}{Whole Heart}
		& FS Baseline & 27.66$\pm$10.88 & 16.50$\pm$7.19 & 110.74$\pm$20.49 & 34.48$\pm$8.15 \\
		& & SSL Baseline & 35.68$\pm$16.10 & 22.86$\pm$11.79 & 20.73$\pm$21.25 & 8.46$\pm$7.99 \\
		& & SemiSAM+ & 34.24$\pm$20.63 & 20.78$\pm$10.19 & 37.21$\pm$31.35 & 13.92$\pm$11.87 \\
		& & SemiSAM-O1 (R1) & 50.21$\pm$9.28 & 34.09$\pm$9.15 & 89.42$\pm$38.32 & 30.85$\pm$12.58 \\
		& & SemiSAM-O1 (R2) & 65.03$\pm$5.79 & 48.46$\pm$6.50 & 78.46$\pm$35.14 & 22.43$\pm$10.01 \\
		& & SemiSAM-O1 (R3) & \textbf{68.40$\pm$4.95} & \textbf{52.20$\pm$5.86} & \textbf{45.59$\pm$31.72} & \textbf{12.30$\pm$5.60} \\ \hline
		
		\multirow{12}{*}{RT-EC}
		& \multirow{6}{*}{Clinical Target Volume}
		& FS Baseline & 20.05$\pm$12.09 & 11.73$\pm$8.68 & 212.58$\pm$48.26 & 104.76$\pm$33.78 \\
		& & SSL Baseline & 19.39$\pm$12.18 & 11.31$\pm$8.48 & 202.17$\pm$37.28 & 108.37$\pm$32.39 \\
		& & SemiSAM+ & 28.96$\pm$13.92 & 17.76$\pm$10.22 & 202.93$\pm$60.54 & 85.64$\pm$29.60 \\
		& & SemiSAM-O1 (R1) & 36.43$\pm$16.99 & 23.49$\pm$11.78 & 30.95$\pm$18.27 & 12.32$\pm$13.79 \\
		& & SemiSAM-O1 (R2) & 42.36$\pm$16.23 & 28.11$\pm$12.08 & 25.01$\pm$21.27 & \textbf{7.14$\pm$11.10} \\
		& & SemiSAM-O1 (R3) & \textbf{44.13$\pm$16.00} & \textbf{29.53$\pm$11.98} & \textbf{24.72$\pm$21.40} & 7.19$\pm$10.97 \\ \cline{2-7}

& \multirow{6}{*}{Planning Target Volume}
& FS Baseline & 24.36$\pm$9.43 & 14.19$\pm$6.03 & 28.95$\pm$8.21 & 8.63$\pm$6.01 \\
& & SSL Baseline & 27.42$\pm$10.51 & 16.30$\pm$6.89 & 28.81$\pm$8.85 & 6.85$\pm$3.84 \\
& & SemiSAM+ & 44.93$\pm$15.97 & 30.28$\pm$12.67 & 47.80$\pm$44.36 & 16.63$\pm$13.22 \\
& & SemiSAM-O1 (R1) & 42.00$\pm$7.44 & 26.89$\pm$6.56 & 30.90$\pm$6.02 & 16.36$\pm$3.85 \\
& & SemiSAM-O1 (R2) & 43.78$\pm$7.29 & 28.33$\pm$6.53 & 42.34$\pm$42.22 & 18.15$\pm$11.32 \\
& & SemiSAM-O1 (R3) & \textbf{47.33$\pm$5.94} & \textbf{31.22$\pm$5.43} & \textbf{28.36$\pm$8.44} & \textbf{14.05$\pm$4.61} \\ \hline \hline
	\end{tabular}
\end{table*}

\begin{figure}[t]
\centering
\includegraphics[width=\linewidth]{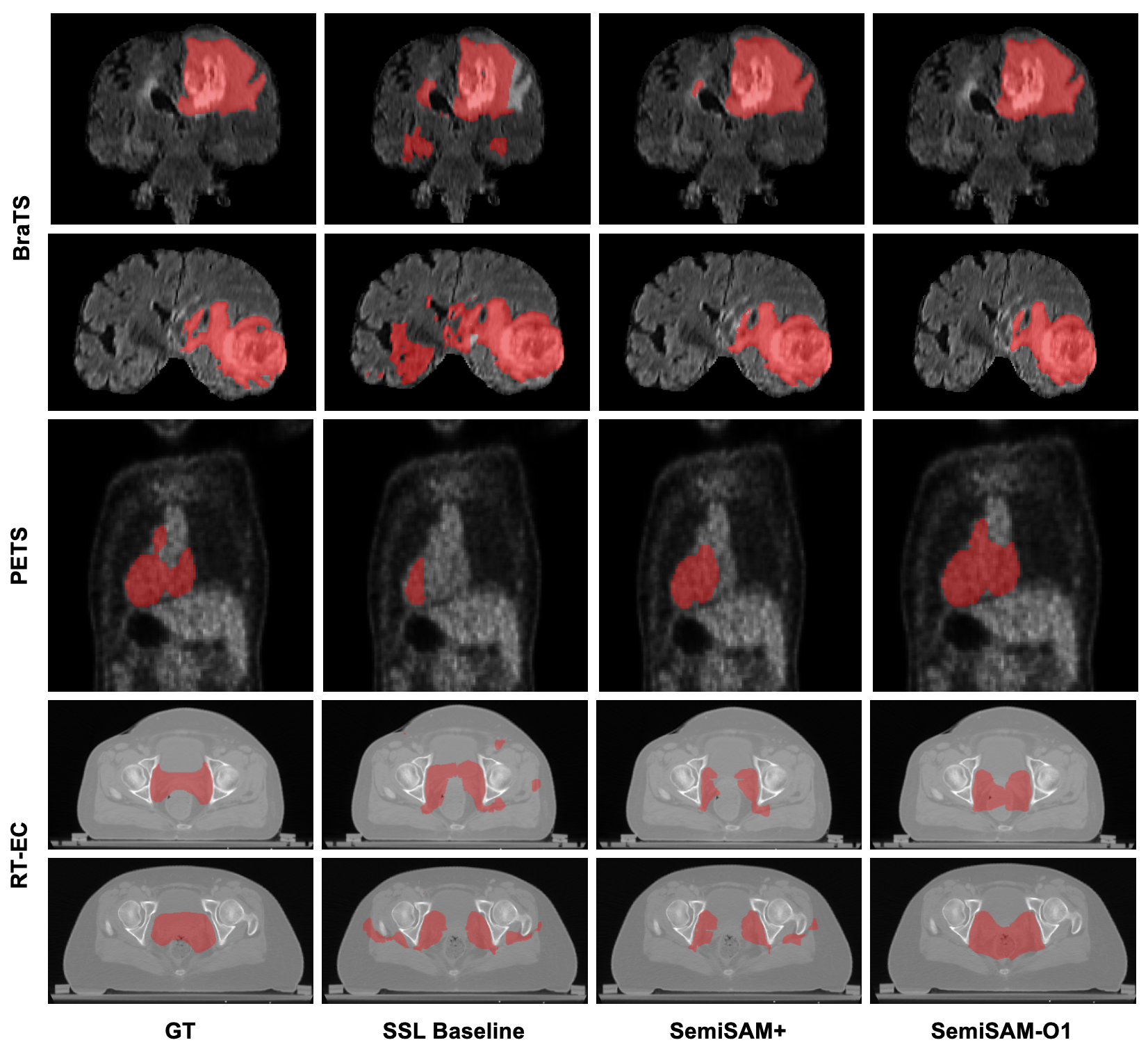}
\caption{Qualitative comparison of segmentation results across different segmentation tasks across different modalities and anatomical under the one-label setting.}
\label{fig_visall}
\end{figure}

\subsection{Evaluation on a Wide Range of Segmentation Tasks}

In addition to LA, we further conduct extensive cross-dataset evaluations to verify the generalizability and robustness of SemiSAM-O1 across diverse segmentation tasks, covering different imaging modalities, anatomical targets, and clinical scenarios.
All experiments adopt the MT backbone with 3D U-Net as the specialist model and SAM-Med3D as the generalist model, with quantitative results and qualitative visualization summarized in Table~\ref{Table_cross} and Figure~\ref{fig_visall}.

On the BraTS 2019 dataset for brain tumor segmentation, the high morphological heterogeneity, irregular boundaries, and large inter-patient variability of brain tumors pose significant challenges.
It can be observed that the FS baseline achieves a poor Dice of 42.82\%, and the SSL baseline yields a limited improvement to 52.56\%.
SemiSAM+ boosts the performance to 68.53\% Dice by leveraging foundation model-driven consistency regularization.
It is worth noting that SemiSAM-O1 (R1) achieves a Dice of only 61.60\%, falling below SemiSAM+. 
We attribute this to the fact that the high inter-patient heterogeneity in brain tumor morphology, size, and location prevents a single template prototype from adequately covering the entire feature distribution space. 
As iterative refinement progressively corrects these initialization errors, SemiSAM-O1 achieves comparable segmentation accuracy (68.50\% Dice at R3) while delivering notably superior boundary delineation, with 95HD reduced from 38.81 to 33.97 and ASD from 16.66 to 14.26. This indicates that, beyond improving coarse region overlap, the proposed iterative refinement primarily enhances structural consistency and boundary precision. Importantly, SemiSAM-O1 achieves this without invoking the repeated calls of SAM inference during training, resulting in significantly lower training cost compared with SemiSAM+.

For the PETS dataset, the pre-trained SAM-Med3D foundation model was exclusively trained on anatomical modalities and has never encountered functional PET data during pre-training, resulting in a substantial domain gap. In this scenario, SemiSAM+ suffers from performance degradation, achieving a Dice of only 34.24\%, even lower than the vanilla SSL baseline (35.68\%). This indicates that the online consistency regularization strategy is highly vulnerable to generalization failure of the foundation model on unseen modalities.
In stark contrast, SemiSAM-O1 effectively exploits the foundation model's feature representation capability even under this domain gap. Through iterative pseudo-label refinement, SemiSAM-O1 progressively improves from 50.21\% (R1) to 68.40\% (R3), substantially surpassing both fully supervised and semi-supervised methods.

The RT-EC dataset presents a considerably more challenging scenario owing to the inherently ambiguous and diffuse boundaries of CTV and PTV in radiotherapy planning.
The FS baseline achieves only 20.05\% Dice for CTV and 24.36\% Dice for PTV, highlighting the difficulty of learning reliable target representations from a single labeled volume.
For CTV, the SSL baseline further degrades performance to 19.39\%, even falling below the FS baseline, while only providing a marginal improvement for PTV. This behavior exposes a critical limitation of conventional SSL approaches under extreme low-data regimes: although the FS baseline benefits from clean supervision on the labeled sample, the MT framework introduces noisy consistency constraints when the teacher model is poorly initialized, which can hinder optimization instead of improving generalization.
In contrast, SemiSAM-O1 effectively overcomes this representational collapse by leveraging the foundation model as a strong prior for pseudo-label initialization and iterative refinement. Across successive refinement rounds, SemiSAM-O1 consistently improves segmentation performance for both target structures, achieving 44.13\% Dice on CTV and 47.33\% Dice on PTV at R3. Notably, these consistent gains across two anatomically distinct radiotherapy targets demonstrate that SemiSAM-O1 remains highly effective in multi-target segmentation. Nevertheless, the absolute performance remains relatively modest compared with other datasets, likely because SAM-Med3D was not explicitly pre-trained on radiotherapy target delineation. This observation further suggests that relying on a single generalist foundation model may not be universally optimal for highly specialized medical domains, motivating future foundation models with stronger representations.

\section{Discussion and Conclusion}

In this work, we investigate the extreme low-annotation regime for medical image segmentation and push the boundary of semi-supervised learning to the one-label setting. We propose SemiSAM-O1, a novel framework that effectively leverages foundation model representations to enable high-quality segmentation using only a single annotated volume. Through a carefully designed iterative pseudo-label refinement pipeline, the proposed method establishes a strong specialist–generalist collaboration without relying on extensive manual annotations or heavy online prompting.

The key insight of this work lies in rethinking the role of foundation models beyond promptable segmentation. Instead of treating the generalist model solely as an inference-time assistant, SemiSAM-O1 fully exploits its feature space as a global semantic embedding to guide both initialization and refinement of pseudo-labels. The prototype-based propagation provides a meaningful starting point under extreme supervision scarcity, while the uncertainty-guided KNN refinement introduces a robust mechanism to correct errors by leveraging inter-sample relationships. This design transforms the learning process into a progressively improving cycle, where pseudo-label quality and model capability are mutually enhanced.
Extensive experiments across multiple datasets, modalities, and SSL backbones demonstrate that SemiSAM-O1 consistently achieves substantial performance gains over existing semi-supervised and SAM-based fine-tuning methods. Notably, the performance gap between one-label learning and full supervision is significantly reduced, suggesting that reliable segmentation models can be trained with minimal annotation effort. In addition, the proposed framework improves computational efficiency by decoupling the generalist model from the training loop, avoiding the expensive online inference required by prior methods.

Despite these promising results, several limitations remain. The performance of SemiSAM-O1 still depends on the quality and generalization ability of the adopted foundation model. Any inherent bias or domain gap in the learned feature space may affect both the initialization and subsequent refinement of pseudo-labels.
Moreover, our current study is limited to pre-defined generalist models, which may not consistently perform well across all tasks and imaging modalities due to variations in modality characteristics, anatomical structures, and data distributions \citep{zhang2026uncovering}. This suggests that relying on a single generalist model may inherently constrain the upper bound of performance.
A promising direction for future work is to explore more flexible and adaptive utilization of multiple foundation models \citep{zou2025fusionfm}.
Incorporating modality-specific foundation models may provide more discriminative representations for certain tasks \citep{wittmann2025vesselfm,zhang2025seganypet,zhang2026developing,marks2025cellsam}.
Building upon this idea, one potential solution is to construct a pool of generalist foundation models and dynamically select or combine the most suitable model(s) for a given task. This could be further extended to an agentic framework, where different models are treated as agents that collaboratively contribute to the learning procedure \citep{huang2026medsegagent}. Such a framework has the potential to mitigate model-specific biases and significantly enhance robustness and generalization across diverse clinical scenarios.

Besides, our approach assumes that semantically similar cases retrieved in the foundation-model feature space exhibit sufficient structural consistency for voxel-wise pseudo-label aggregation.
Nevertheless, this assumption becomes less reliable for anatomies with extreme inter-patient variability, such as highly heterogeneous tumors, where semantically similar cases may still exhibit considerable local differences in lesion size, shape, and spatial location.  This phenomenon is also reflected in our BraTS experiments, where the first refinement round provides less improvement than on anatomically more consistent datasets.  Future work could relax this assumption by incorporating deformable feature alignment, anatomy-aware correspondence estimation, or local feature matching before pseudo-label aggregation, thereby extending the proposed framework to anatomies with more complex structural variability.

In conclusion, SemiSAM-O1 demonstrates that it is possible to train high-performing medical image segmentation models with only one annotated example by effectively leveraging foundation model representations and unlabeled data. This work highlights a new paradigm for annotation-efficient learning and provides a strong step toward reducing the reliance on costly expert annotations in medical imaging.

\bibliographystyle{model2-names.bst}\biboptions{authoryear}
\bibliography{refs}

\end{document}